\documentclass[lettersize,journal]{IEEEtran}
\usepackage{amsmath,amsfonts}
\usepackage{algpseudocode}
\usepackage{algorithm}
\usepackage{array}

\usepackage{subcaption}

\usepackage{textcomp}
\usepackage{stfloats}
\usepackage{url}
\usepackage{verbatim}
\usepackage{graphicx}
\usepackage{cite}
\usepackage{booktabs}
\usepackage{adjustbox}
\usepackage{colortbl}
\usepackage{multirow}

\usepackage{amssymb}
\usepackage{xcolor}

\usepackage{pifont}   

\usepackage{tabularx}
\usepackage{makecell}
\usepackage[colorlinks=true,linkcolor=blue,citecolor=blue,urlcolor=blue,bookmarks=false]{hyperref}

\usepackage[normalem]{ulem}

\definecolor{tabfirst}{rgb}{1, 0.7, 0.7}
\definecolor{tabsecond}{rgb}{1, 0.85, 0.7}
\definecolor{tabthird}{rgb}{1, 1, 0.7}

\newcommand{\first}{\cellcolor{tabfirst}}
\newcommand{\second}{\cellcolor{tabsecond}}
\newcommand{\third}{\cellcolor{tabthird}}  

\newcommand{\firstbox}[1]{\colorbox{tabfirst}{#1}}
\newcommand{\secondbox}[1]{\colorbox{tabsecond}{#1}}
\newcommand{\thirdbox}[1]{\colorbox{tabthird}{#1}}

\usepackage[font=small,labelfont=bf,tableposition=top]{caption}

\begin{document}
\title{DrawSplat: Atlanta World-Guided Implicit Gaussian SLAM for Indoor Reconstruction}
	
\author{Fan Zhu, Zhisong Xu, Ziyu Chen, Zhenjun Zhao, Takayuki Okatani
	\thanks{This work was supported by JST BOOST, Japan Grant Number JPMJBS2421. (Corresponding author: Takayuki Okatani.)}%
	\thanks{Fan Zhu and Takayuki Okatani are with the Graduate School of Information Sciences, Tohoku University, Sendai, Japan.}
	\thanks{Takayuki Okatani is also with the RIKEN Center for AIP, Tokyo, Japan.}
	\thanks{Zhisong Xu is with the Department of Electrical Engineering and Information Systems, Graduate School of Engineering, The University of Tokyo, Tokyo, Japan.}
	\thanks{Ziyu Chen is with the Hefei Institutes of Physical Science, Chinese Academy of Sciences, and Department of Automation, University of Science and Technology of China, Hefei, China.}
	\thanks{Zhenjun Zhao is with the Department of Computer and Systems Engineering, University of Zaragoza, Zaragoza, Spain.}
}
	
\maketitle



\begin{abstract}
	RGB-D SLAM requires accurate online tracking, high-quality rendering, and geometrically consistent reconstruction. Existing Gaussian SLAM methods mainly optimize Gaussian primitives with photometric, depth, or local frame-to-model constraints, which are often insufficient in indoor scenes with weak textures, large planar structures, noisy depth, and long trajectories. The resulting maps may suffer from pose drift, surface thickness artifacts, noisy normals, and blurred structural boundaries. We propose DrawSplat, an Atlanta World-guided implicit Gaussian SLAM framework for indoor reconstruction. DrawSplat follows a drawing-like process: explicit point-line geometry first sketches the structural layout, and implicit neural Gaussians then fill in high-fidelity appearance and continuous geometric details. It estimates camera poses, sparse geometry, and gravity direction through point-line joint tracking,  infers structural planes from tracked lines using the shared-gravity component of the Atlanta World prior, and converts them into differentiable regularization terms on Gaussian centers and normals. To improve compactness and efficiency, DrawSplat further combines an anchor-based implicit neural Gaussian representation with adaptive structural sampling, which allocates Gaussian capacity to structurally informative regions. Experiments on multiple datasets show that DrawSplat achieves robust tracking, high-fidelity rendering, and strong geometric reconstruction among Gaussian-based SLAM systems, while maintaining a practical balance among quality, efficiency, and model size.
\end{abstract}

\begin{IEEEkeywords}
Gaussian Splatting, Visual SLAM, Reconstruction, RGB-D, Atlanta World.
\end{IEEEkeywords}

\section{Introduction}

\begin{figure}[!tpb]
	\centering
	\includegraphics[width=0.95\linewidth]{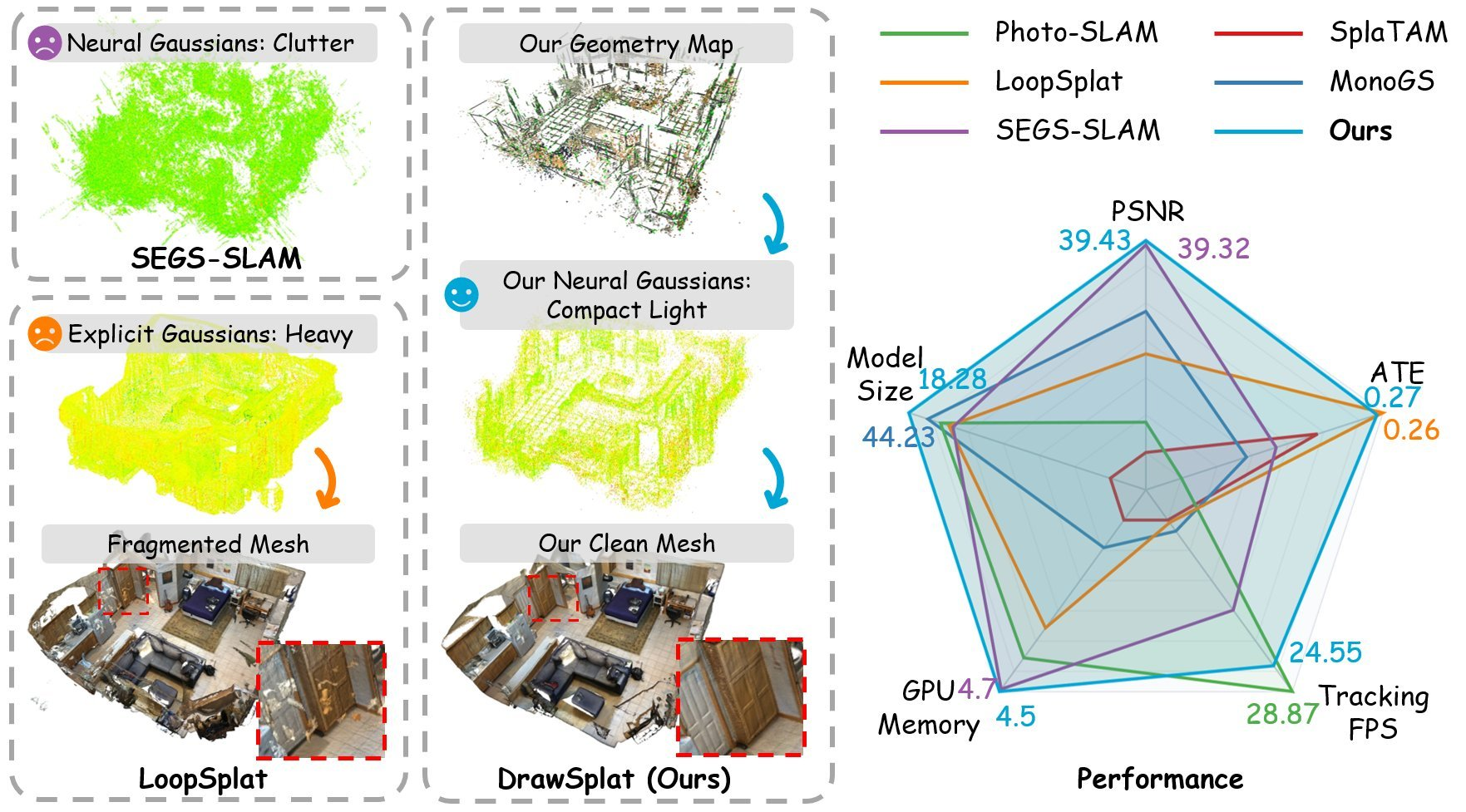}
	\caption{Compared with the cluttered neural Gaussian map of SEGS-SLAM~\cite{Segs-slam} and the heavy explicit Gaussian map of LoopSplat~\cite{loopsplat}, DrawSplat achieves a compact map and a clean mesh by using the Atlanta World assumption to guide implicit neural Gaussians. The right-side radar chart is normalized by using the best and worst values of each metric as boundaries, and shows the relative performance of our method in terms of tracking, rendering, real-time capability, and cost metrics on the Replica~\cite{replica} dataset.}
	\label{fig:motivation}
\end{figure}
\IEEEPARstart{V}{isual} simultaneous localization and mapping (SLAM) aims to estimate camera motion and reconstruct 3D scenes from online image streams~\cite{ORB3,SSMB,AdaptVPR}. Beyond accurate tracking, modern SLAM systems are expected to build photorealistic, geometrically accurate, and structurally consistent scene representations~\cite{advances3dv,meshaligned,framevggt}, supporting free-viewpoint rendering and reliable geometric reasoning for augmented reality, robotics, embodied scene understanding, and spatial intelligence~\cite{lrsgs,EAI_OV,passlam,PanoImager,FoL++}.

Neural radiance field (NeRF)~\cite{Nerf}-based SLAM improves dense reconstruction through continuous implicit representations~\cite{Nice-slam,Co-slam,Eslam,spslam,skanslam}, but its optimization and rendering costs remain high for online use~\cite{GaussR-SLAM}. In contrast, 3D Gaussian Splatting~\cite{3dgs} and some variations~\cite{scaffoldgs,2dgs,GP-GS} provide efficient differentiable rendering~\cite{howsurvey}, motivating recent Gaussian SLAM systems for online mapping and novel-view synthesis~\cite{Splatam,MonoGS,CG-SLAM,Rtg-slam,loopsplat,MG-SLAM,mygosplat}.

Despite this progress, Gaussian SLAM still struggles to build geometrically reliable indoor maps under weak textures, large planar regions, occlusions, noisy depth, and long trajectories~\cite{MG-SLAM,AtlasGS}. In such cases, photometric, depth, or local frame-to-model constraints alone may be insufficient, while indoor scenes provide useful structural regularities such as vertical-horizontal consistency, wall-floor intersections, planar boundaries, and long-range layout coherence. Without such guidance, Gaussian maps may suffer from pose drift, surface thickness artifacts, noisy normals, blurred boundaries, or incomplete planar reconstruction in corridors, staircases, walls, and door frames~\cite{scaffoldgs,Segs-slam}. Existing structural or implicit-prior methods~\cite{MG-SLAM,Densesplat} alleviate these issues, but often introduce additional overhead, structural redundancy, or difficulty in maintaining appearance-geometry consistency.

Our core idea is to make Gaussian SLAM follow a \textbf{drawing-like} process: \textit{explicit geometry first sketches the scene structure, and implicit Gaussians then recover appearance and local details.} In indoor environments, point-line geometry provides a sparse but stable structural sketch, including geometric skeletons, gravity direction, and structural planes, to guide differentiable Gaussian optimization while keeping the map compact.

To realize this idea, we propose DrawSplat, an Atlanta World~\cite{AWA}-guided implicit Gaussian SLAM framework for indoor reconstruction. DrawSplat follows a hybrid Gaussian SLAM paradigm~\cite{HI-SLAM2}: the point-line front-end estimates poses, refines sparse geometry, and infers gravity without relying on the evolving Gaussian appearance map, while the back-end incrementally builds and optimizes an Atlanta World-guided implicit Gaussian map. Tracked line structures and gravity further induce  structural planes constrained by the shared-gravity Atlanta World prior to suppress depth noise and enhance structural regularity. Finally, adaptive structural sampling and a joint map optimization objective enable efficient, compact, and appearance-geometry-consistent map generation, as shown in Fig.~\ref{fig:motivation}.

We evaluate DrawSplat on multiple datasets. It achieves strong tracking accuracy and reconstruction quality among Gaussian-based SLAM systems while maintaining a practical quality-efficiency trade-off. Compared with the second-best baseline, DrawSplat reduces trajectory error by 60\% on the multi-room Replica Apartment dataset and improves rendering quality by 2.54 dB on the real-world ScanNet dataset. The main contributions are:
\begin{itemize}
	\item We propose an Atlanta World-guided Gaussian SLAM framework that combines explicit geometric tracking with implicit neural Gaussian mapping for high-fidelity indoor reconstruction with appearance-geometry consistency.
	\item We introduce an  Atlanta World-guided structural regularization strategy that converts tracked lines, gravity cues, and inferred structural planes into differentiable constraints on Gaussian primitives.
	\item We design an adaptive structural sampling strategy that reallocates Gaussian capacity to structurally informative regions, improving the balance between reconstruction quality and runtime efficiency.
\end{itemize}

\section{Related Works}

\begin{figure*}[!tpb]
	\centering
	\includegraphics[width=1\linewidth]{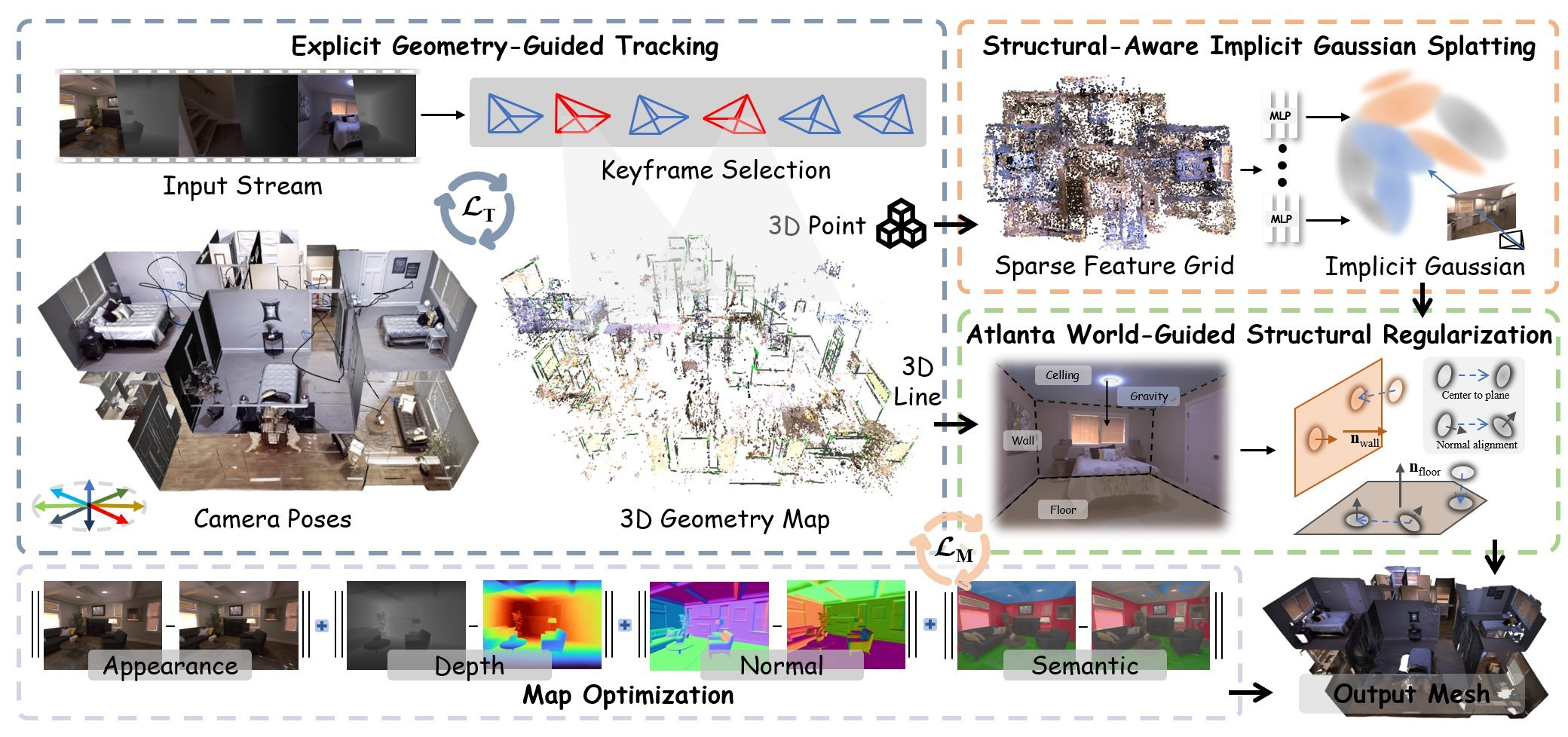}
	\caption{Overview of DrawSplat. {DrawSplat follows a hybrid Gaussian SLAM paradigm~\cite{HI-SLAM2}: explicit point-line geometry provides stable tracking and structural cues, while the implicit Gaussian map is incrementally updated online for compact, photorealistic, and geometry-consistent reconstruction.} Given an online RGB-D stream, the tracking front-end jointly estimates camera poses, selects keyframes, and builds a sparse 3D geometry map from point and line features. The mapping back-end converts tracked 3D points into a sparse feature grid and decodes implicit Gaussian primitives through lightweight MLPs. Structural 3D lines further induce  structural planes constrained by the shared-gravity Atlanta World prior, which regularize Gaussian centers and normals during joint map optimization. The final system produces compact, photorealistic, and structurally consistent indoor reconstructions.}
	\label{fig:overview}
\end{figure*}

\subsection{Neural Implicit SLAM}

Unlike traditional visual SLAM systems~\cite{ElasticFusion, Orb-slam2, ORB3} that represent scenes with point clouds, voxels, or TSDF volumes~\cite{balf,SIDslam}, neural implicit SLAM methods use continuous scene representations for dense reconstruction~\cite{Nice-slam,Co-slam,Eslam,spslam,skanslam}. Starting from iMAP~\cite{Imap} and NICE-SLAM~\cite{Nice-slam}, later methods improve efficiency and geometry with multi-scale feature grids, tri-plane features, hash grids, or neural point clouds~\cite{Vox-fusion,Co-slam,Eslam,Point-slam}. These representations often recover smooth surfaces, but dense field queries, volumetric rendering, and iterative optimization limit real-time performance and interactive rendering~\cite{howsurvey}. DrawSplat instead keeps the efficient differentiable rasterization of 3D Gaussian Splatting~\cite{3dgs}, while using anchor-based implicit Gaussian decoding to introduce continuity and structural compatibility for online mapping.

\subsection{Gaussian-based SLAM}

3D Gaussian Splatting~\cite{3dgs} has been widely used in SLAM for online mapping and photorealistic rendering~\cite{SemanticGaussians,GraphGSOcc,gssurvey,mmdslam,RP-SLAM}. SplaTAM~\cite{Splatam} and MonoGS~\cite{MonoGS} directly optimize Gaussian primitives for tracking and reconstruction, while Photo-SLAM~\cite{Photo-slam} and RTG-SLAM~\cite{Rtg-slam} combine feature-based tracking with Gaussian mapping for efficient pose estimation. Splat-SLAM~\cite{splat-slam} and LoopSplat~\cite{loopsplat} further address global optimization, loop constraints, and long-sequence consistency, with extensions to dynamic~\cite{dygs,GARAD-SLAM,dgsslam} and outdoor scenes~\cite{OutdoorMS,Wildgs-slam}.

Despite these advances, accurate geometry remains difficult because explicit Gaussian primitives are discrete and are often optimized from local photometric or depth cues. For Gaussian reconstruction, GP-GS~\cite{GP-GS} uses Gaussian-process regression and uncertainty filtering to densify sparse SfM point clouds for initialization. DrawSplat instead allocates Gaussian capacity during incremental mapping using line, depth-gradient, and photometric responses. Methods based on 3DGS variants~\cite{FGO-SLAM, FGO-SLAM++, HI-SLAM2, gof, Radegs} improve surface quality, and some systems introduce monocular estimation modules to avoid noisy captured depth~\cite{HI-SLAM2, splat-slam, IMGS-SLAM}. Other methods~\cite{Mgs-slam,dmslam,EGC-SLAM} use multi-view stereo (MVS)~\cite{mvs,mvs2} or 3D foundation models~\cite{Dust3r,mast3r} to obtain geometric information from monocular input. These designs may increase complexity or underuse the structural information available in RGB-D SLAM~\cite{f2slam}. DrawSplat adopts a hybrid Atlanta World-guided Gaussian SLAM design that uses explicit geometry for stable tracking and structural cues, and implicit Gaussians for compact, high-quality mapping.

\subsection{Structural Priors for Gaussian SLAM}

Man-made indoor scenes contain structural regularities that have long been used to improve SLAM stability. Traditional structured SLAM methods exploit lines, planes, and Manhattan World constraints for pose estimation and map consistency in weak-texture scenes~\cite{PLVS,ManhattanSLAM}. Atlanta World relaxes the single Manhattan frame by allowing multiple horizontal dominant directions while preserving a shared gravity direction, making it better suited to multi-room indoor layouts~\cite{AWA}.

MG-SLAM~\cite{MG-SLAM} is a representative Manhattan World-based Gaussian SLAM method. It uses line and planar constraints for tracking and mesh completion~\cite{manhattan}, but its structural priors are not fully integrated into differentiable Gaussian map optimization. DenseSplat~\cite{Densesplat} instead uses a NeRF prior to densify Gaussian maps under sparse keyframes and occlusions, improving continuity at the cost of an additional prior model.  AtlasGS~\cite{AtlasGS} studies Atlanta World-guided implicit Gaussian reconstruction with precomputed poses, whereas DrawSplat derives structural planes from online point-line tracking and updates them during mapping. Compared with SEGS-SLAM~\cite{Segs-slam}, DrawSplat introduces line-induced plane regularization and adaptive structural sampling into incremental neural Gaussian mapping.

\section{Methods}

Given an RGB-D image stream $\mathcal{I}=\{(\mathbf{C}_t,\mathbf{D}_t)\}_{t=1}^{T}$, our goal is to reconstruct indoor scenes that are simultaneously photorealistic, geometrically accurate, and structurally consistent. To this end, we propose a unified framework that couples explicit geometric tracking with a structure-aware implicit Gaussian representation. Specifically, the system first organizes keyframe geometry into an anchor-based implicit Gaussian map (Sec.~\ref{sec:implicit_gaussian}). It then estimates camera poses, a sparse geometric map, and the global gravity direction through point-line joint tracking (Sec.~\ref{sec:tracking}). Based on these geometric cues, we further introduce an Atlanta World (AW) prior that converts explicit structural planes into differentiable constraints on Gaussian primitives (Sec.~\ref{sec:aw_regularization}). Finally, we improve reconstruction quality through adaptive structural sampling and joint map optimization while preserving practical online efficiency (Sec.~\ref{sec:mapoptimization}). Fig.~\ref{fig:overview} provides an overview of our pipeline.

\subsection{Structure-Aware Implicit Gaussian Splatting}
\label{sec:implicit_gaussian}
We represent the scene with a structure-aware implicit Gaussian formulation inspired by~\cite{scaffoldgs, AtlasGS} rather than a set of unconstrained explicit 3D Gaussians~\cite{3dgs, 2dgs, gof}. The key idea is to use sparse anchor points as the spatial backbone and decode local neural Gaussian primitives around each anchor. This design preserves the rendering efficiency of Gaussian Splatting while introducing the spatial continuity and structural compatibility needed for downstream regularization.

\subsubsection{Incremental Voxel Feature Grid Construction}
\label{sec:voxel_grid}
In online SLAM, the scene continuously expands as new keyframes arrive. The scene representation must therefore support dynamic growth without relying on a globally precomputed point cloud. Similar to~\cite{Segs-slam}, for the point cloud of keyframe $k$, denoted by $\mathcal{P}_k=\{\mathbf{p}_{k,j}\}_{j=1}^{N_k}\subset\mathbb{R}^{3}$, we discretize space into voxels with resolution $\epsilon$ and construct an anchor set $\mathcal{V}_k$. Each point $\mathbf{p}_{k,j}$ is mapped to the base coordinate of its voxel:

\begin{equation}
\mathbf{v}_{k,j}=\epsilon\left\lfloor \frac{\mathbf{p}_{k,j}}{\epsilon}\right\rfloor .
\label{eq:voxel_anchor}
\end{equation}

This yields the occupied anchor set

\begin{equation}
\mathcal{V}_k=\left\{\mathbf{v}_{k,j}\mid \mathbf{p}_{k,j}\in \mathcal{P}_k\right\}.
\label{eq:anchor_set}
\end{equation}

This incremental voxelization suppresses redundant points while retaining the main geometric structure of the scene. {The anchor grid is updated incrementally rather than rebuilt from scratch. When refined keyframe geometry is fed back to the voxelization process, points falling into already occupied voxels update the corresponding anchors, while points in newly occupied voxels create new anchors. Existing anchors are preserved unless they are later removed by the structure-aware pruning strategy in Sec.~\ref{sec:adaptive_sampling}.}

Following bundle adjustment, pose corrections are propagated to existing anchors and their local Gaussian geometry through anchor--keyframe associations. Gaussian centers undergo the corresponding rigid transformations, while local displacements and orientations receive the rotational corrections. Anchor identities and learned features are retained, and the map is further refined through subsequent mapping optimization.

Each anchor $\mathbf{v}\in\mathcal{V}_k$ serves not only as a spatial index but also as a local representation carrier. Specifically, we associate each anchor with a feature vector $\mathbf{f}_v\in\mathbb{R}^{32}$, a local scale factor $l_v\in\mathbb{R}$, and a set of learnable offsets $\{\mathbf{o}_{v,i}\}_{i=1}^{K}$ used to generate $K$ local Gaussian primitives. 

For a camera center $\mathbf{x}_c$, we compute the scalar anchor-to-camera distance
$\rho_{vc}=\left\lVert\mathbf{v}-\mathbf{x}_c\right\rVert_2$ and the unit viewing-direction vector $\mathbf{d}_{vc}=(\mathbf{v}-\mathbf{x}_c)/{\rho_{vc}}.$ Both quantities are used as conditioning signals for Gaussian decoding.

Anchors provide stable spatial organization, while decoded local Gaussians supply fine-detail capacity, allowing the representation to grow online without the redundancy of purely explicit Gaussian maps.

\subsubsection{Neural Gaussian Parameter Decoding}
\label{sec:gaussian_decoding}
Anchors alone cannot be rendered directly. We therefore need to decode them into Gaussian primitives with geometry- and appearance-related attributes, while keeping the representation compatible with later structural constraints. For each visible anchor $\mathbf{v}$, we first compute the center of the $i$-th Gaussian primitive as

\begin{equation}
\boldsymbol{\mu}_{v,i}=\mathbf{v}+l_v\mathbf{o}_{v,i}.
\label{eq:gaussian_center}
\end{equation}

This step expands the discrete anchor skeleton into a continuous set of local Gaussian elements that cover the nearby spatial region. We then use lightweight MLPs to decode the remaining primitive attributes. Let $\boldsymbol{\theta}_{v,i}$ denote the attribute vector of the $i$-th primitive generated from anchor $\mathbf{v}$. We define

\begin{equation}
\boldsymbol{\theta}_{v,i}=
\mathcal{M}_{a}\left(\mathbf{f}_v,\rho_{vc},\mathbf{d}_{vc},\mathbf{o}_{v,i}\right),
\label{eq:gaussian_attribute}
\end{equation}
where $\boldsymbol{\theta}_{v,i}$ includes the color $\mathbf{c}_{v,i}$, opacity $\alpha_{v,i}$, rotation $\mathbf{r}_{v,i}$, scale $\mathbf{s}_{v,i}$, and a semantic attribute $\mathbf{e}_{v,i}$ used later for structure-aware weighting. In this way, anchors provide stable local context, while the Gaussian attributes are generated adaptively according to spatial and view-dependent cues.

The combination of sparse anchors and neural decoding achieves both compactness and continuity: only anchors are stored explicitly, while primitive attributes are generated as continuous functions for stable reconstruction in low-texture and planar regions.

\subsubsection{Differentiable Multi-modal Rendering}
\label{sec:multi_modal_rendering}

Once the 3D Gaussian parameters are obtained, we render color, depth, normal, and semantic attributes through a unified differentiable renderer. For a pixel $\mathbf{x}'$, a rendered attribute $\mathbf{A}(\mathbf{x}')$ is computed by depth-ordered alpha compositing:

\begin{equation}
\mathbf{A}(\mathbf{x}')=
\sum_{i\in\mathcal{N}(\mathbf{x}')}
\mathbf{a}_i w_i
\prod_{j=1}^{i-1}\left(1-w_j\right).
\label{eq:alpha_compositing}
\end{equation}

where $w_i=\alpha_i\mathcal{G}'_i(\mathbf{x}')$ is the contribution weight of the $i$-th Gaussian at the pixel, $\mathcal{G}'_i$ is its projected distribution on the image plane, and $\mathbf{a}_i$ denotes the corresponding rendered attribute.

This unified renderer lets photometric, depth, semantic, and structural supervision operate on the same Gaussian primitives, providing a single differentiable path for updating the scene representation.

\subsection{Explicit Geometry-Guided Tracking}
\label{sec:tracking}

Relying only on an implicit scene representation is insufficient for stable pose estimation in low-texture indoor environments. Inaccurate poses would, in turn, degrade Gaussian initialization and subsequent map optimization. We therefore introduce an explicit geometry-guided tracking front-end that provides reliable camera poses, sparse scene geometry, and gravity cues for the implicit Gaussian back-end.

\subsubsection{Point-Line Feature Extraction and Association}

{Our tracking front-end is built upon the well-established point-based SLAM method~\cite{ORB3}, which provides reliable local correspondences and efficient camera pose estimation. Unlike Gaussian SLAM systems that mainly rely on point features or feature-based tracking for pose estimation~\cite{Photo-slam,Segs-slam}, DrawSplat further extracts line features from each RGB-D frame to expose structural cues that are important for indoor reconstruction and subsequent Gaussian regularization.}

{Specifically, for each RGB-D frame, we extract ORB point features~\cite{ORB3} and detect line segments using EDLines~\cite{EDLines} over an image pyramid. Inter-frame point associations are established using ORB descriptors, while line associations are established using LBD descriptors. Point features provide conventional local correspondence constraints in textured regions, whereas line features remain stable along wall boundaries, door frames, floor boundaries, and ceiling contours. Their joint use improves tracking robustness in weak-texture indoor regions and provides explicit structural observations for gravity estimation, plane inference, and structure-aware Gaussian optimization.}

\subsubsection{Point-Line Joint Pose Estimation}
\label{sec:point_line_pose}

Given point and line correspondences, we estimate the current camera pose $\mathbf{T}_t=(\mathbf{R}_t,\mathbf{t}_t)\in \mathrm{SE}(3)$ by minimizing multi-source reprojection and geometric residuals. For a map point $\mathbf{P}_m\in\mathcal{P}$ and its image observation $\mathbf{u}_m=[u_m,v_m]^\top$, the 2D point reprojection residual is defined as

\begin{equation}
\mathbf{e}_{p,m}^{\mathrm{2D}}=
\mathbf{u}_m-\pi\left(\mathbf{R}_t\mathbf{P}_m+\mathbf{t}_t\right).
\label{eq:point_reprojection}
\end{equation}

For a map line segment $\mathbf{L}_n=(\mathbf{P}_n^b,\mathbf{P}_n^e)\in\mathcal{X}_{\ell}$ and its associated image line $\ell_n$, we measure the distances from the projected endpoints to the observed image line. Let $\mathbf{l}_n$ be the normalized homogeneous line vector of $\ell_n$, and let $\tilde{\pi}(\mathbf{x})=[\pi(\mathbf{x})^\top,1]^\top$ denote the homogeneous image projection. The 2D line residual is

\begin{equation}
\mathbf{e}_{\ell,n}^{\mathrm{2D}}=
\begin{bmatrix}
\mathbf{l}_n^\top\tilde{\pi}\left(\mathbf{R}_t\mathbf{P}_n^b+\mathbf{t}_t\right)\\
\mathbf{l}_n^\top\tilde{\pi}\left(\mathbf{R}_t\mathbf{P}_n^e+\mathbf{t}_t\right)
\end{bmatrix}.
\label{eq:line_reprojection}
\end{equation}

The 2D tracking loss is then written as

\begin{equation}
\begin{aligned}
\mathcal{L}^{\mathrm{2D}}
&=
\mathcal{L}_{p}^{\mathrm{2D}}+
\mathcal{L}_{\ell}^{\mathrm{2D}}\\
&=
\sum_m \rho\left(\left\lVert\mathbf{e}_{p,m}^{\mathrm{2D}}\right\rVert_{\boldsymbol{\Omega}_{p,m}^{\mathrm{2D}}}^{2}\right)
+
\sum_n \rho\left(\left\lVert\mathbf{e}_{\ell,n}^{\mathrm{2D}}\right\rVert_{\boldsymbol{\Omega}_{\ell,n}^{\mathrm{2D}}}^{2}\right),
\end{aligned}
\label{eq:tracking_2d_loss}
\end{equation}
where $\boldsymbol{\Omega}_{p,m}^{\mathrm{2D}}$ and $\boldsymbol{\Omega}_{\ell,n}^{\mathrm{2D}}$ are information matrices that encode detection uncertainty, and $\rho(\cdot)$ is a robust kernel.

Since RGB-D input is available, we further back-project observed points and line endpoints into 3D. For the observed point $\mathbf{u}_m$, its measured 3D position is

\begin{equation}
\hat{\mathbf{P}}_m=\Psi\left(\mathbf{u}_m,D_t(\mathbf{u}_m)\right),
\label{eq:point_backprojection}
\end{equation}
where $\Psi(\cdot)$ denotes the back-projection function. Similarly, the observed 2D endpoints of line $\ell_n$ are back-projected to obtain a measured 3D segment $\hat{\mathbf{L}}_n=(\hat{\mathbf{P}}_n^b,\hat{\mathbf{P}}_n^e)$. We then define the 3D point and line residuals as

\begin{align}
\mathbf{e}_{p,m}^{\mathrm{3D}}
&=
\hat{\mathbf{P}}_m-
\left(\mathbf{R}_t\mathbf{P}_m+\mathbf{t}_t\right),
\label{eq:point_3d_residual}\\
\mathbf{e}_{\ell,n}^{\mathrm{3D}}
&=
\begin{bmatrix}
d\left(\mathbf{R}_t\mathbf{P}_n^b+\mathbf{t}_t,\hat{\mathbf{L}}_n\right)\\
d\left(\mathbf{R}_t\mathbf{P}_n^e+\mathbf{t}_t,\hat{\mathbf{L}}_n\right)
\end{bmatrix},
\label{eq:line_3d_residual}
\end{align}
where $d(\mathbf{x},\hat{\mathbf{L}}_n)$ denotes the Euclidean distance from point $\mathbf{x}$ to the measured 3D line segment $\hat{\mathbf{L}}_n$. The 3D geometric loss is

\begin{equation}
\begin{aligned}
\mathcal{L}^{\mathrm{3D}}
&=
\mathcal{L}_{p}^{\mathrm{3D}}+
\mathcal{L}_{\ell}^{\mathrm{3D}}\\
&=
\sum_m \rho\left(\left\lVert\mathbf{e}_{p,m}^{\mathrm{3D}}\right\rVert_{\boldsymbol{\Omega}_{p,m}^{\mathrm{3D}}}^{2}\right)
+
\sum_n \rho\left(\left\lVert\mathbf{e}_{\ell,n}^{\mathrm{3D}}\right\rVert_{\boldsymbol{\Omega}_{\ell,n}^{\mathrm{3D}}}^{2}\right).
\end{aligned}
\label{eq:tracking_3d_loss}
\end{equation}

The overall point-line tracking objective combines the 2D reprojection and 3D geometric terms:

\begin{equation}
\mathcal{L}_{\mathrm{T}}=
\lambda_{p}^{\mathrm{2D}}\mathcal{L}_{p}^{\mathrm{2D}}+
\lambda_{\ell}^{\mathrm{2D}}\mathcal{L}_{\ell}^{\mathrm{2D}}+
\lambda_{p}^{\mathrm{3D}}\mathcal{L}_{p}^{\mathrm{3D}}+
\lambda_{\ell}^{\mathrm{3D}}\mathcal{L}_{\ell}^{\mathrm{3D}}.
\label{eq:tracking_loss}
\end{equation}

The robust Huber cost function $\rho\left(\cdot\right)$ in Eqs.~\eqref{eq:tracking_2d_loss} and~\eqref{eq:tracking_3d_loss} suppresses the impact of mismatches and outliers during optimization.

Compared with point-only tracking, the joint point-line formulation provides stronger geometric pose locking. This is particularly important in corridors, stairs, and large planar regions, where line constraints remain informative even when local texture is weak.

\subsubsection{Local and Global Geometric Refinement}

To reduce long-term drift, we further refine the front-end estimates through local and global geometric optimization. Local bundle adjustment optimizes the current keyframe, its covisible keyframes, and the associated sparse geometric map. When a loop is detected, a BoW~\cite{BoW}-based retrieval module triggers global bundle adjustment to jointly refine the entire camera trajectory and the explicit geometric map. The optimized 3D geometry is then fed back into the incremental voxelization process of Sec.~\ref{sec:implicit_gaussian} to update the anchor grid.

The optimized 3D geometry provides globally correctable trajectories and sparse structural cues for the implicit Gaussian back-end.

\subsubsection{Gravity Estimation from Structural Lines}
\label{sec:gravity_estimation}

Indoor environments usually obey strong vertical-horizontal structural regularities. A stable gravity direction is essential for the subsequent AW-based structural modeling.

Let $\mathcal{X}_{\ell}=\{\mathbf{L}_i\}_{i=1}^{N_l}$ denote the tracked 3D line set, with unit direction

\begin{equation}
\mathbf{d}_i=
\frac{\mathbf{P}_i^e-\mathbf{P}_i^b}
{\left\lVert\mathbf{P}_i^e-\mathbf{P}_i^b\right\rVert_2}.
\label{eq:line_direction}
\end{equation}

Under the AW assumption, structural lines are expected to be either parallel or orthogonal to the unit gravity direction $\mathbf{n}_g$. We estimate gravity by minimizing

\begin{align}
\mathcal{L}_{g}(\mathbf{n})
&=\sum_{i=1}^{N_l} \omega_i
\left(
\min\left(
|\mathbf{d}_i^{\top}\mathbf{n}|,
1-|\mathbf{d}_i^{\top}\mathbf{n}|
\right)
\right)\\
\mathbf{n}_g
&=\operatorname*{arg\,min}_{\left\lVert\mathbf{n}\right\rVert_2=1}
\mathcal{L}_{g}(\mathbf{n}).
\label{eq:gravity_argmin}
\end{align}

{The reliability weight $\omega_i=\omega_i^{\mathrm{geo}}\omega_i^{\mathrm{sem}}$ combines geometric reliability and semantic structural confidence.
Here, the \(\omega_i^{\mathrm{geo}}\) is computed from line length and observation frequency, and \(\omega_i^{\mathrm{sem}}\) is obtained from the semantic probabilities along the projected line segment. Lines classified as structural elements, such as wall, floor, or ceiling boundaries, receive higher weights, while non-structural or diagonal objects are down-weighted.}

Because indoor line structures often correspond to stable elements such as wall corners, door frames, and ceiling boundaries, line-based gravity estimation is generally more robust than point-cloud-based alternatives in structured indoor scenes.

\subsection{ Atlanta World-Guided Structural Regularization}
\label{sec:aw_regularization}

Accurate poses and continuous Gaussian rendering do not automatically guarantee structurally correct surfaces. In low-texture planar regions, photometric and depth supervision alone often lead to thickness artifacts, noisy normals, and local geometric drift.

To address this issue, we introduce an Atlanta World-guided planar formulation that uses the gravity direction to regularize horizontal and vertical structural surfaces. We integrate this prior with tracked line-induced planes to guide online implicit Gaussian mapping.
 
The overall logic can be summarized as $\mathcal{X}_{\ell}\rightarrow \Pi \rightarrow \mathcal{G}$,
where $\mathcal{X}_{\ell}=\{\mathbf{L}_i\}$ denotes the 3D line set, $\Pi=\{\Pi_j\}$ denotes the inferred structural plane set, and $\mathcal{G}=\{G_k\}$ denotes the Gaussian primitives regularized by those planes.

\subsubsection{Line-induced Plane Inference}
\label{sec:plane_inference}

Compared with sparse points, line features provide stronger directional regularity and are better suited for recovering indoor structural planes.

Let $\mathcal{X}_{\ell}=\{\mathbf{L}_i\}_{i=1}^{N_l}$ be the tracked line set, and let $\mathbf{d}_i$ denote the unit direction of each line. We aim to infer a plane set $\Pi=\{\Pi_j\}_{j=1}^{N_\pi}$, where each plane is parameterized as $\Pi_j=(\mathbf{n}_j,\delta_j)$  with unit normal $\mathbf{n}_j$ ($\|\mathbf{n}_j\|_2=1$) and offset $\delta_j$.

We define three types of geometric constraints. First, direction consistency enforces that a line direction lies in its assigned plane:

\begin{equation}
e_{\mathrm{dir}}^{ij}=\left(\mathbf{n}_j^\top \mathbf{d}_i\right)^2.
\label{eq:plane_direction_residual}
\end{equation}

Second, position consistency enforces that the line endpoints lie on the plane:

\begin{equation}
e_{\mathrm{pos}}^{ij}=
\left(\mathbf{n}_j^\top \mathbf{P}_i^b+\delta_j\right)^2+
\left(\mathbf{n}_j^\top \mathbf{P}_i^e+\delta_j\right)^2.
\label{eq:plane_position_residual}
\end{equation}

Third, AW consistency requires the plane normal to be either parallel or orthogonal to gravity:

\begin{equation}
e_{\mathrm{aw}}^{j}=
\min\left(
\left|\mathbf{n}_j^\top \mathbf{n}_g\right|,
1-\left|\mathbf{n}_j^\top \mathbf{n}_g\right|
\right).
\label{eq:plane_aw_residual}
\end{equation}
Since $\mathbf{n}_j$ and $\mathbf{n}_g$ are unit vectors, the absolute inner product lies in $[0,1]$. The residual is therefore non-negative and invariant to reversing either normal, with zero error for parallel or orthogonal directions.

The final plane inference objective is

\begin{equation}
\mathcal{L}_{\mathrm{plane}}=
\sum_{i,j} z_{ij}
\left(e_{\mathrm{dir}}^{ij}+\lambda_{\mathrm{pos}}e_{\mathrm{pos}}^{ij}\right)
+\lambda_{\mathrm{paw}}\sum_j e_{\mathrm{aw}}^{j}.
\label{eq:plane_loss}
\end{equation}
where $z_{ij}\in\{0,1\}$ indicates whether line $\mathbf{L}_i$ is assigned to plane $\Pi_j$, and $\lambda_{\mathrm{pos}}$ and $\lambda_{\mathrm{paw}}$ balance the position and plane-level AW terms.
Each plane has an independently estimated unit normal and offset, and the shared-gravity prior constrains its inclination. The variables $z_{ij}$ assign lines to fitted planes, and they do not assign planes to a discrete set of local Manhattan frames.

When a new keyframe arrives, newly reconstructed 3D lines are first associated with existing planes. RANSAC initializes additional planes from unexplained line sets that satisfy the minimum line-support and semantic-compatibility requirements, with inliers identified by directional and point-to-plane residuals. Existing planes retain their identities, while their unit normals and offsets are refined during subsequent mapping using the line-fitting and structural objectives. Gaussian primitives are associated only with geometrically and semantically compatible planes.
Sparse local line observations are converted into structural plane targets for Gaussian regularization.

\subsubsection{Semantic-Assisted Plane Weighting}
\label{sec:semantic_plane_weighting}

Not every region in an indoor scene strictly follows the AW assumption. Furniture, decorations, and clutter often coexist with structural walls, floors, and ceilings. Applying the same plane regularization everywhere would over-smooth non-structural objects.

We therefore introduce soft plane-to-Gaussian association weights. When semantic labels are available, we further classify planes into categories such as wall, floor, ceiling, and others, and use this information to modulate regularization strength. For a Gaussian primitive $G_k$ centered at $\mathbf{P}_k$, we first compute its point-to-plane Euclidean distance

\begin{equation}
d_{kj}=\left|\mathbf{n}_j^\top\mathbf{P}_k+\delta_j\right|.
\label{eq:point_plane_distance}
\end{equation}

We then define a normalized soft weight

\begin{equation}
w_{kj}=
\frac{\exp\left(-\eta d_{kj}\right)s_{kj}}
{\sum_m \exp\left(-\eta d_{km}\right)s_{km}}.
\label{eq:plane_gaussian_weight}
\end{equation}

where $s_{kj}\in[0,1]$ denotes the semantic compatibility between Gaussian $G_k$ and the category $c_j$ of plane $\Pi_j$. It is obtained from the semantic probability associated with $G_k$ and is high when the Gaussian and the plane have compatible structural categories. Gaussians classified as non-structural regions are assigned zero structural association, and $\eta$ is a distance decay factor.

The normalization is applied only over semantically compatible structural planes; Gaussian primitives without a valid structural association receive zero structural weight.
This formulation combines spatial proximity with semantic plausibility. Structural planes therefore exert stronger constraints on nearby and semantically compatible Gaussians, while non-structural objects retain greater geometric flexibility.

\subsubsection{Plane-to-Gaussian Regularization}
\label{sec:plane_gaussian_regularization}

Once the soft plane-to-Gaussian associations are established, we translate structural priors into differentiable losses on Gaussian parameters. Let the covariance of Gaussian $G_k$ be decomposed as $\Sigma_k=\mathbf{R}_k\Lambda_k\mathbf{R}_k^\top$, and let $\mathbf{n}_k$ be the local Gaussian normal defined by the eigenvector corresponding to the smallest eigenvalue.

We first define a position regularization term that pulls Gaussian centers toward their associated structural planes:

\begin{equation}
\mathcal{L}_{\mathrm{gp}}=
\sum_{k,j} w_{kj}
\left(\mathbf{n}_j^\top \mathbf{P}_k+\delta_j\right)^2.
\label{eq:gaussian_position_loss}
\end{equation}

We then define a normal regularization term that aligns Gaussian normals with plane normals:

\begin{equation}
\mathcal{L}_{\mathrm{gn}}=
\sum_{k,j} w_{kj}
\left(1-\left|\mathbf{n}_k^\top \mathbf{n}_j\right|\right).
\label{eq:gaussian_normal_loss}
\end{equation}
The absolute inner product makes this normal alignment term invariant to the arbitrary orientation of either normal.

Finally, we impose AW consistency directly on the structural planes:

\begin{equation}
\mathcal{L}_{\mathrm{aw}}=
\sum_j
\begin{cases}
\left|\mathbf{n}_j^\top \mathbf{n}_g\right|, & c_j=\mathrm{wall},\\
1-\left|\mathbf{n}_j^\top \mathbf{n}_g\right|, &
c_j\in\{\mathrm{floor},\mathrm{ceiling}\},\\
0, & c_j=\mathrm{others}.
\end{cases}
\label{eq:aw_loss}
\end{equation}

The unified structural loss is defined as

\begin{equation}
\mathcal{L}_{\mathrm{struct}}=
\lambda_{\mathrm{gp}}\mathcal{L}_{\mathrm{gp}}+
\lambda_{\mathrm{gn}}\mathcal{L}_{\mathrm{gn}}+
\lambda_{\mathrm{aw}}\mathcal{L}_{\mathrm{aw}}.
\label{eq:structural_loss}
\end{equation}

Minimizing $\mathcal{L}_{\mathrm{struct}}$ during map optimization suppresses planar thickness artifacts, stabilizes local normals, and preserves global vertical-horizontal consistency.

This transfers explicit structural geometry into Gaussian optimization, especially benefiting large planar and weakly textured indoor regions.

\begin{algorithm}[t]
	\caption{Adaptive Structural Sampling}
	\label{alg:adaptive_sampling}
	\begin{algorithmic}[1]
		\Require Candidate offset centers $\mathcal{X}$, anchor set $\mathcal{A}$, tracked 3D lines $\mathcal{L}$, rendered image $\hat{I}$, rendered depth $\hat{D}$, input image $I$
		\Ensure Updated anchor set $\mathcal{A}$
		\For{each candidate center $\mathbf{x}\in\mathcal{X}$}
		\State Project $\mathbf{x}$ to the current keyframe as $\mathbf{u}$
		\State Compute $\rho_{\ell}(\mathbf{x})$ from $\mathcal{L}$ using Eq.~\eqref{eq:line_response}
		\State Compute $\rho_{d}(\mathbf{x})$ and $\rho_{c}(\mathbf{x})$ using Eqs.~\eqref{eq:depth_response}--\eqref{eq:color_response}
		\State Normalize the three responses within the current keyframe
		\State Compute $w(\mathbf{x})$ using Eq.~\eqref{eq:structural_weight}
		\State Compute $s(\mathbf{x})$ using Eq.~\eqref{eq:densification_score}
		\If{$s(\mathbf{x})>\tau_s$}
		\State Spawn a new anchor at the corresponding multi-scale voxel level
		\State Initialize its feature from the parent anchor and set local offsets to zero
		\EndIf
		\EndFor
		\For{each anchor $a\in\mathcal{A}$}
		\State Compute $\bar{w}_a=\frac{1}{K}\sum_{j=1}^{K}w(\mathbf{x}_{a,j})$
		\If{$n_a>\tau_n$ and $\max_j\alpha_{a,j}<\tau_{\alpha}$ and $\bar{w}_a<\tau_w$}
		\State Remove anchor $a$
		\EndIf
		\EndFor
	\end{algorithmic}
\end{algorithm}

\subsection{Map Optimization}
\label{sec:mapoptimization}

\begin{figure*}[!tpb]
	\centering
	\includegraphics[width=0.98\linewidth]{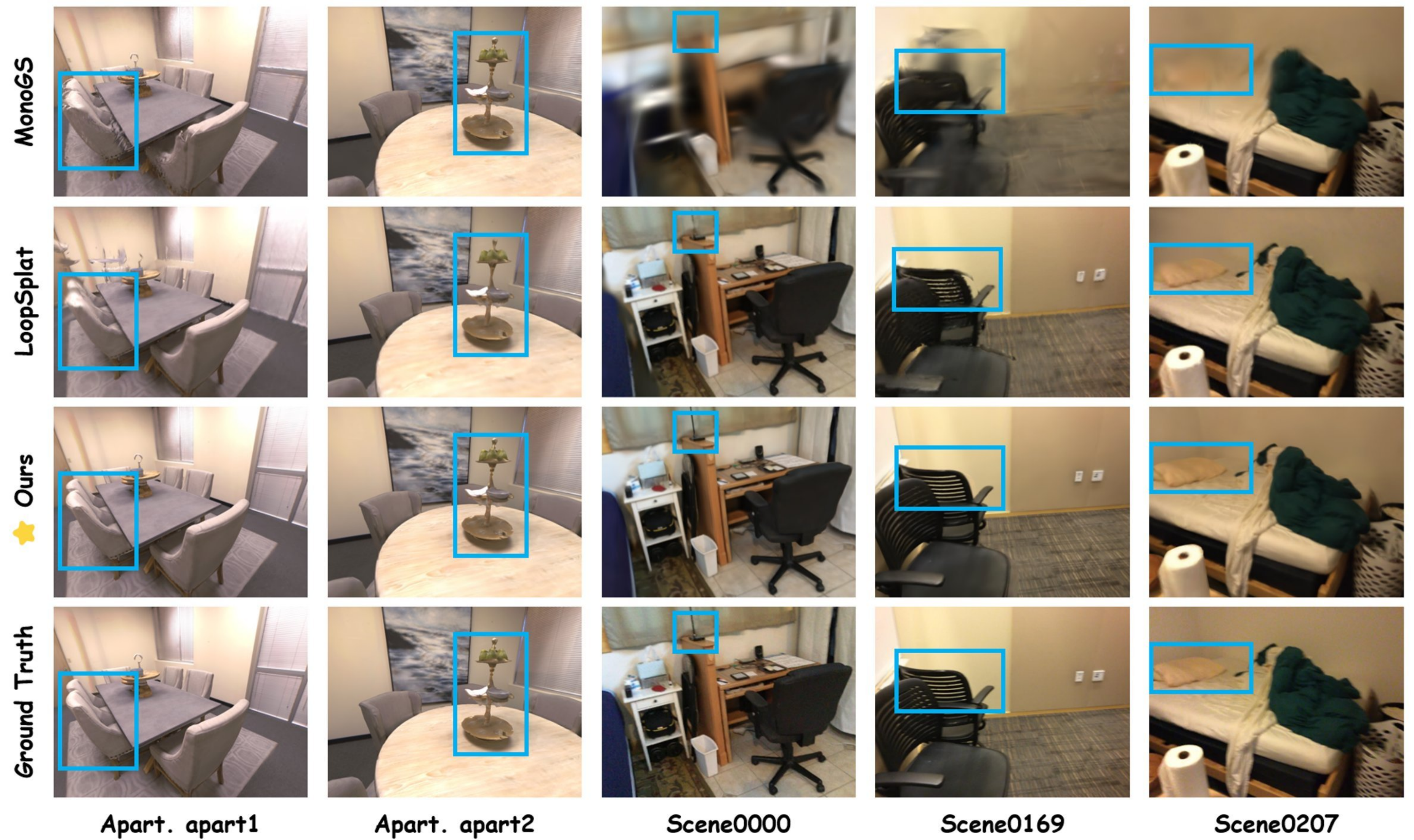}
	\caption{Rendering comparison on the Replica Apartment~\cite{replica} and ScanNet~\cite{Scannet} datasets. Our method preserves finer visual details.}
	\label{fig:rendering}
\end{figure*}

\subsubsection{Adaptive Structural Sampling}
\label{sec:adaptive_sampling}
Uniform densification wastes model capacity on simple regions and under-represents difficult geometric areas such as corners, door frames, and depth discontinuities. We therefore introduce adaptive structural sampling, which allocates Gaussian capacity according to explicit line structure, depth discontinuity, and photometric residual.

For an arbitrary spatial location $\mathbf{x}$, we first define a line response using the tracked 3D line set:

\begin{equation}
\rho_{\ell}(\mathbf{x})=
\sum_i \exp\left(
-\frac{d^2(\mathbf{x},\ell_i)}{\sigma_{\ell}^{2}}
\right).
\label{eq:line_response}
\end{equation}
where $d(\mathbf{x},\ell_i)$ is the shortest distance from $\mathbf{x}$ to the 3D line segment $\ell_i$, and $\sigma_{\ell}$ controls the spatial support radius. This term increases the sampling priority near wall intersections, door frames, and structural boundaries.

To capture geometric discontinuities on visible surfaces, we further introduce a depth-gradient response and a photometric residual response:

\begin{align}
\rho_{d}(\mathbf{x})
&=\left\lVert\nabla \hat{D}(\mathbf{u})\right\rVert_2, \label{eq:depth_response}\\
\rho_{c}(\mathbf{x})
&=\left\lVert I(\mathbf{u})-\hat{I}(\mathbf{u})\right\rVert_1.
\label{eq:color_response}
\end{align}
where $\mathbf{u}$ denotes the projection of $\mathbf{x}$ into the current keyframe. After normalization, the three responses are fused into a unified structural weight

\begin{equation}
w(\mathbf{x})=
\alpha_{\ell}\rho_{\ell}(\mathbf{x})+
\alpha_{d}\rho_{d}(\mathbf{x})+
\alpha_{c}\rho_{c}(\mathbf{x}).
\label{eq:structural_weight}
\end{equation}

Rather than replacing the original gradient-based densification criterion, we use the structural weight to modulate it. Let $g(\mathbf{x})$ denote the accumulated screen-space gradient magnitude of the offset Gaussian centered at $\mathbf{x}$. We define the final densification score as

\begin{equation}
s(\mathbf{x})=
\left\lVert g(\mathbf{x})\right\rVert_2
\left(1+w(\mathbf{x})\right).
\label{eq:densification_score}
\end{equation}

When the score of an offset is high, the system spawns a new anchor at the corresponding multi-scale voxel level. The new anchor inherits the parent feature context, while its local offsets are initialized to zero.

During topology simplification, we use conservative structure-aware pruning. For an anchor with $K$ offsets, we compute the average structural weight

\begin{equation}
\bar{w}=\frac{1}{K}\sum_{j=1}^{K}w(\mathbf{x}_j).
\label{eq:average_structural_weight}
\end{equation}

An anchor is removed only if it has been sufficiently observed, all its offsets have low opacity, and its average structural weight is below the threshold, i.e., $\bar{w}<\tau_w$. The complete procedure is summarized in Algorithm~\ref{alg:adaptive_sampling}.

Adaptive structural sampling improves the quality-efficiency trade-off by reallocating Gaussians toward structurally informative regions rather than uniformly increasing map density.

\subsubsection{Joint Map Optimization Objective}
\label{sec:map_objective}

The final map optimization objective combines appearance, depth, normal, structural, semantic, and distribution regularization:

\begin{equation}
\mathcal{L}_{\mathrm{M}}=
\mathcal{L}_{\mathrm{rgb}}+
\lambda_{d}\mathcal{L}_{\mathrm{d}}+
\lambda_{n}\mathcal{L}_{\mathrm{n}}+
\lambda_{s}\mathcal{L}_{\mathrm{struct}}+
\lambda_{\mathrm{sem}}\mathcal{L}_{\mathrm{sem}}+
\lambda_{o}\mathcal{L}_{\mathrm{dist}}.
\label{eq:map_loss}
\end{equation}
where the photometric loss is defined as

\begin{equation}
\mathcal{L}_{\mathrm{rgb}}=\sum_{\mathbf{u}}
\left\lVert C(\mathbf{u})-\hat{C}(\mathbf{u})\right\rVert_1.
\label{eq:rgb_loss}
\end{equation}

The depth loss is evaluated over the valid-depth pixel set $\Omega_D$:
	
\begin{equation}
	\mathcal{L}_{\mathrm{d}}=
	\sum_{\mathbf{u}\in\Omega_D}
	\left\lVert D(\mathbf{u})-\hat{D}(\mathbf{u})\right\rVert_1.
	\label{eq:depth_loss}
\end{equation}

The semantic loss is computed as a pixel-wise cross-entropy between the observed semantic label $S_c(\mathbf{u})$ and rendered semantic probability $\hat{S}_c(\mathbf{u})$:

\begin{equation}
\mathcal{L}_{\mathrm{sem}}=
-\sum_{\mathbf{u}}\sum_c
S_c(\mathbf{u})\log \hat{S}_c(\mathbf{u}).
\label{eq:semantic_loss}
\end{equation}

To further improve surface smoothness and suppress high-frequency geometric noise, we introduce a normal consistency term:

\begin{equation}
\mathcal{L}_{\mathrm{n}}=
\sum_{\mathbf{u}\in\Omega_D}
\left(1-\mathbf{n}(\mathbf{u})^\top\hat{\mathbf{n}}(\mathbf{u})\right).
\label{eq:normal_loss}
\end{equation}
where $\mathbf{n}(\mathbf{u})$ represents the rendered normal and $\hat{\mathbf{n}}(\mathbf{u})$ represents the normal calculated from the input depth map.

In Eq.~\eqref{eq:map_loss}, $\mathcal{L}_{\mathrm{struct}}$ is given by Eq.~\eqref{eq:structural_loss} and injects explicit structural priors into the optimization, while $\mathcal{L}_{\mathrm{dist}}$, adapted from~\cite{2dgs}, regularizes the primitive distribution to maintain rendering stability. By jointly minimizing these objectives, the system improves planar regularity, boundary sharpness, and global topological consistency while preserving rendering quality.

This joint objective unifies appearance fidelity and geometric consistency in a single optimization framework. Appearance terms preserve rendering fidelity, geometric terms stabilize local surfaces, structural terms enforce global regularity, and adaptive sampling allocates optimization capacity to structurally informative regions.

For clarity, let $\Pi=\{\Pi_j\}_{j=1}^{N_\pi}$ denote the set of structural planes, where $\Pi_j=(\mathbf{n}_j,\delta_j)$, and let $\boldsymbol{\Theta}_{G}$ denote the learnable parameters of the implicit Gaussian map, including anchor features, local scales, offsets, and decoder parameters. Let $\mathcal{Z}=\{z_{ij}\}$ denote the line-to-plane assignments defined in Eq.~\eqref{eq:plane_loss}. Within each mapping batch, the current camera poses $\{\mathbf{T}_t\}$ and gravity direction $\mathbf{n}_g$ are fixed, and the plane parameters and Gaussian map parameters are updated alternately. For compactness, fixed observations and loss arguments are omitted below:

\begin{equation}
	\begin{aligned}
		\Pi &\leftarrow \operatorname*{arg\,min}_{\Pi:\,\|\mathbf{n}_j\|_2=1,\ \forall j}
		\left[\mathcal{L}_{\mathrm{plane}}+\lambda_s\mathcal{L}_{\mathrm{struct}}\right],\\
		\boldsymbol{\Theta}_{G}
		&\leftarrow \operatorname*{arg\,min}_{\boldsymbol{\Theta}_{G}}\mathcal{L}_{\mathrm{M}}.
	\end{aligned}
	\label{eq:alternating_optimization}
	\end{equation}

Here, each $\operatorname*{arg\,min}$ denotes one or a few gradient-based updates rather than full convergence. After each plane update, the plane normals are renormalized to satisfy $\|\mathbf{n}_j\|_2=1$. The line-to-plane assignments $\mathcal{Z}$ and the plane-to-Gaussian weights $w_{kj}$ are refreshed between mapping batches. The $\mathcal{L}_{\mathrm{aw}}$ term therefore constrains the plane normals during the plane update, while the updated plane geometry is transferred to the Gaussian map through $\mathcal{L}_{\mathrm{gp}}$ and $\mathcal{L}_{\mathrm{gn}}$ in the subsequent map update.

\begin{figure*}[!tpb]
	\centering
	\includegraphics[width=1\linewidth]{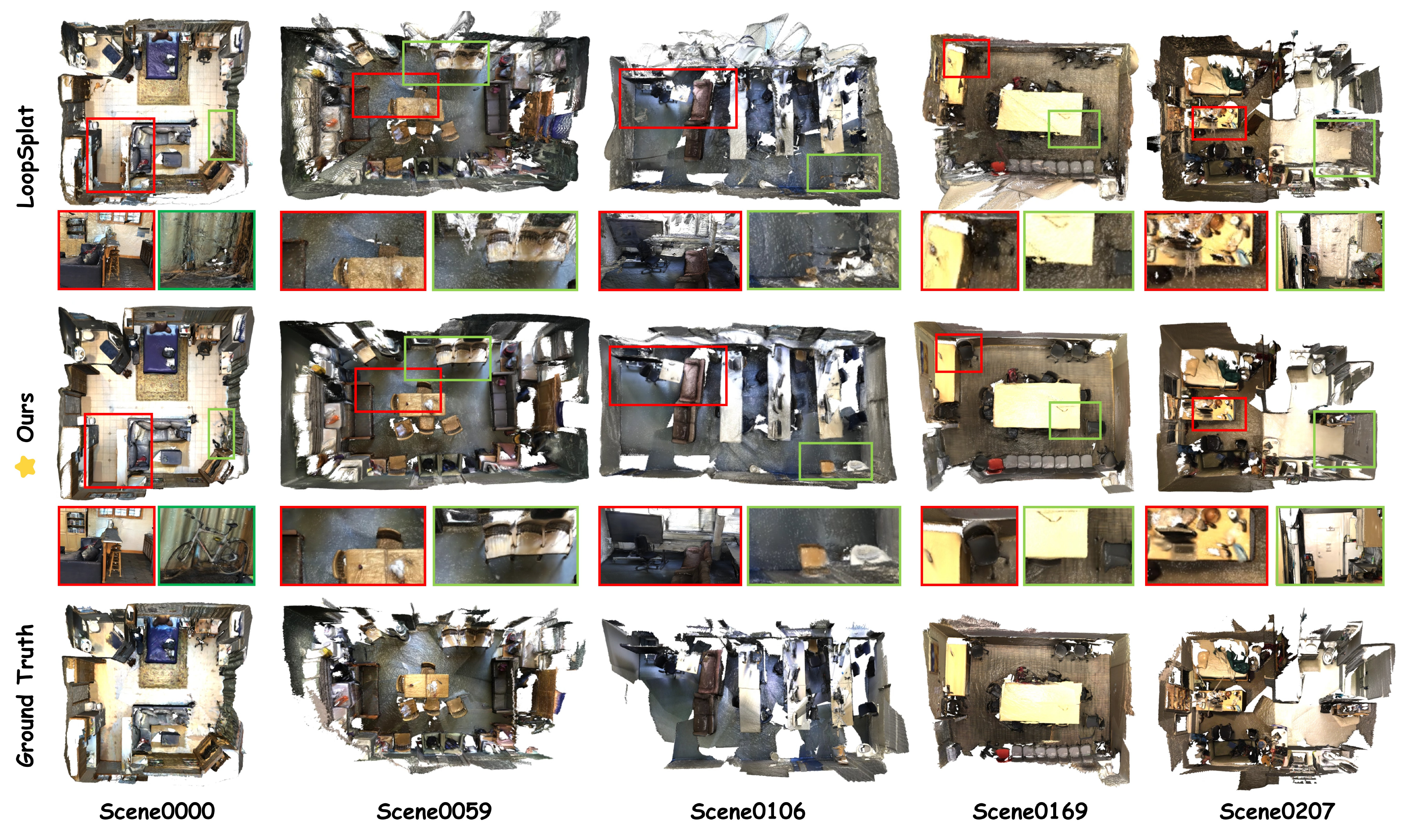}
	\caption{Qualitative geometry reconstruction comparison between our method and representative baselines on the ScanNet dataset~\cite{Scannet}. Our method better preserves scene completeness, structural accuracy, and fine details. The highlighted regions show reconstruction of non-structural objects or locally irregular geometry within otherwise structured indoor scenes, including bicycles, chairs, and table clutter.}
	\label{fig:mesh_scannet}
\end{figure*}

\begin{table}[t]
	\centering
	\caption{Quantitative comparison of camera tracking accuracy (RMSE [cm] $\downarrow$) on the simulated Replica dataset~\cite{replica}.}
	\label{tab:tracking_replica}
	\renewcommand{\arraystretch}{0.8}
	\setlength{\tabcolsep}{3pt}
	\resizebox{0.498\textwidth}{!}{
	\begin{tabular}{lccccccccc}
		\toprule
		\textbf{Method} & Ro0 & Ro1 & Ro2 & Off0 & Off1 & Off2 & Off3 & Off4 & \textbf{Avg.} \\
		\midrule
		SplaTAM~\cite{Splatam}     & 0.31 & 0.47 & 0.26 & 0.43 & \third 0.24 & \third 0.28 & 0.37 & 0.58 & 0.37 \\
		Photo-SLAM~\cite{Photo-slam}  & 0.49 & 0.42 & 0.29 & 0.56 & 0.43 & 1.16 & 0.80 & 0.63 & 0.60 \\
		MonoGS~\cite{MonoGS}      & 0.45 & 0.31 & 0.29 & 0.45 & 0.46 & \second 0.25 & \first 0.16 & 1.58 & 0.49 \\
		RTG-SLAM~\cite{Rtg-slam}    & \first 0.22 & \third 0.27 & 0.21 &  0.33 & \first 0.20 & \first 0.22 & \third 0.24 & 0.44 & \second 0.27 \\
		DROID-Splat~\cite{droidsplat} & \third 0.27 & 0.28 & 0.28 & \first 0.19 & 0.25 & 0.43 & 0.38 & \third 0.34 & \third 0.30 \\
		Splat-SLAM~\cite{splat-slam}  & 0.33 & 0.66 & 0.26 & \third 0.30 & 0.37 & 0.45 & 0.41 & 0.54 & 0.41 \\
		LoopSplat~\cite{loopsplat}   & \second 0.26 & \second 0.24 & \first 0.18 & \second 0.23 & \second 0.21 & 0.40 & \second 0.23 & \second 0.33 & \first 0.26 \\
		SEGS-SLAM~\cite{Segs-slam}   & 0.29 & 0.28 & \third 0.20 & 0.49 & 0.36 & 0.99 & 0.46 & 0.49 & 0.44 \\
		\textbf{Ours}        & \second 0.26 & \first 0.20 & \second 0.19 & 0.35 & \third 0.24 & 0.32 & 0.32 & \first 0.25 & \second 0.27 \\
		\bottomrule
	\end{tabular}
	}
\end{table}

\section{Experiments}
\label{sec:exp}
We evaluate DrawSplat in terms of tracking accuracy, rendering fidelity, geometric reconstruction, and practical efficiency, and compare it with recent SLAM baselines across synthetic and real-world indoor datasets.

\begin{table}[t]
	\centering
	\caption{Quantitative comparison of camera tracking accuracy (RMSE [cm] $\downarrow$) on the real-world ScanNet dataset~\cite{Scannet}.}
	\label{tab:tracking_scannet}
	\resizebox{0.498\textwidth}{!}{
		\begin{tabular}{lcccccc}
			\toprule
			\textbf{Method} & 0000 & 0059 & 0106 & 0169 & 0207 & \textbf{Avg.} \\
			\midrule
			SplaTAM~\cite{Splatam}   
			& 12.57 & 10.15 & 17.78 & 12.65 & 7.98 & 12.23 \\
			
			MonoGS~\cite{MonoGS}    
			& 16.02 & 12.89 & 20.18 & 10.55 & 11.35 & 14.20 \\
			
			Photo-SLAM~\cite{Photo-slam}
			& 7.88 & 7.85 & 9.84 & 9.96 & 7.53 & 8.61 \\
			
			RTG-SLAM~\cite{Rtg-slam}  
			& 8.11 & \third{6.75} & 9.03 & 10.24 & 9.33 & 8.69 \\
			
			MG-SLAM*~\cite{MG-SLAM}   
			& {5.95} & \second{6.41} & {8.07} & \first{7.29} & \first{6.14} & \second{6.77} \\
			
			LoopSplat~\cite{loopsplat} 
			& \third 5.85 & 6.77 & \first{7.16} & 9.58 & \third 6.82 & 7.24 \\
			
			SEGS-SLAM~\cite{Segs-slam} 
			& \second{5.31} & 7.08 & \third 7.87 & \third 7.85 & 6.99 & \third{7.02} \\
			
			\textbf{Ours} 
			& \first{5.03} & \first{6.28} & \second{7.58} & \second{7.32} & \second{6.37} & \first{6.52} \\
			
			\bottomrule
		\end{tabular}
	}
\end{table}

\noindent\textbf{Implementation Details} All experiments are conducted on a single NVIDIA A100 GPU with 40 GB memory and an AMD EPYC 7542 CPU. Unless otherwise specified, all reported results are obtained on the same hardware. For fairness, we run all publicly available baselines using their official implementations, and only report published results for methods that cannot be practically reproduced. The tracking module is implemented in C++, while the mapping module is implemented in Python/CUDA. ORB feature extraction follows the dataset-specific configuration adopted in prior SLAM systems~\cite{PLVS, Segs-slam}, and keyframe selection follows~\cite{ORB3}.  Mapping runs in parallel with the front-end and uses Adam for optimization. Each iteration samples a view from the currently available keyframes, with a total mapping budget of 30K iterations per sequence. For the structure-aware implicit Gaussian representation, the voxel size is set to $0.01$, the number of local offsets per anchor is set to $K=10$, and the anchor feature dimension is set to $32$. Semantic information is  precomputed using the pre-trained Mask2Former~\cite{Mask2Former}. After mapping, rendered depth maps and their corresponding estimated camera poses are fused into a TSDF volume, from which a surface mesh is extracted for geometric evaluation. The weighting hyperparameters $\lambda_{\mathrm{gp}}, \lambda_{\mathrm{gn}}, \lambda_{\mathrm{aw}}, \lambda_{d}, \lambda_{n}, \lambda_{s}, \lambda_{\mathrm{sem}}$, and $\lambda_{o}$ in Eqs.~\eqref{eq:structural_loss} and~\eqref{eq:map_loss} are 0.2, 0.2, 0.1, 0.4, 0.1, 1, 1, and 100.

\noindent\textbf{Datasets} We evaluate the proposed method on four indoor benchmarks with increasing difficulty: Replica~\cite{replica}, ScanNet~\cite{Scannet}, ScanNet++~\cite{Scannet++}, and Replica Apartment~\cite{replica,Tandem}. Replica provides high-quality RGB-D observations and complete ground-truth geometry, and is therefore used for controlled evaluation of rendering and reconstruction quality. ScanNet introduces real sensor noise, depth uncertainty, and more complex scene layouts, making it suitable for testing robustness in realistic indoor environments. Replica Apartment is a larger-scale benchmark with multiple rooms, frequent viewpoint switching, and longer trajectories, and is used to evaluate long-range consistency and scalability. For geometry reconstruction under more challenging conditions, we additionally report qualitative comparisons on ScanNet++~\cite{Scannet++} whenever the original method provides corresponding evidence.

\begin{table}[!t]
	\centering
	\caption{Quantitative comparison of camera tracking accuracy (RMSE [cm] $\downarrow$) on the simulated Replica Apartment dataset~\cite{replica, Tandem}.}
	\label{tab:tracking_apart}
	\renewcommand{\arraystretch}{1}
	\setlength{\tabcolsep}{6pt}
	\resizebox{0.498\textwidth}{!}{
		\begin{tabular}{lcccccc}
			\toprule
			\textbf{Method} & apart0 & apart1 & apart2 & frl0 & frl4 & \textbf{Avg.} \\
			\midrule
			SplaTAM~\cite{Splatam}    
			& 25.14 & 15.67 & 6.17 & 2.74 & 3.31 & 10.61 \\
			
			Photo-SLAM~\cite{Photo-slam} 
			& \third{5.39} & \third{7.98} & 6.31 & 1.09 & 1.27 & 4.41 \\
			
			MonoGS~\cite{MonoGS}     
			& 20.69 & 9.51 & 7.57 & 2.98 & 6.14 & 9.38 \\
			
			RTG-SLAM~\cite{Rtg-slam}   
			& 6.99 & 9.09 & 6.90 & 1.33 & 0.98 & \third{5.06} \\
			
			MG-SLAM*~\cite{MG-SLAM}    
			& \second{5.03} & \second{3.87} & \third{3.15} & \third{0.92} & \third{0.88} & \second{2.77} \\
			
			LoopSplat~\cite{loopsplat}  
			& 5.61 & 21.12 & \second{0.64} & \second{0.60} & \second{0.87} & 5.77 \\
			
			\textbf{Ours}        
			& \first{3.21} & \first{1.19} & \first{0.32} & \first{0.33} & \first{0.42} & \first{1.09} \\
			
			\bottomrule
		\end{tabular}
	}
\end{table}

\begin{figure}[!t]
\centering
\includegraphics[width=0.98\linewidth]{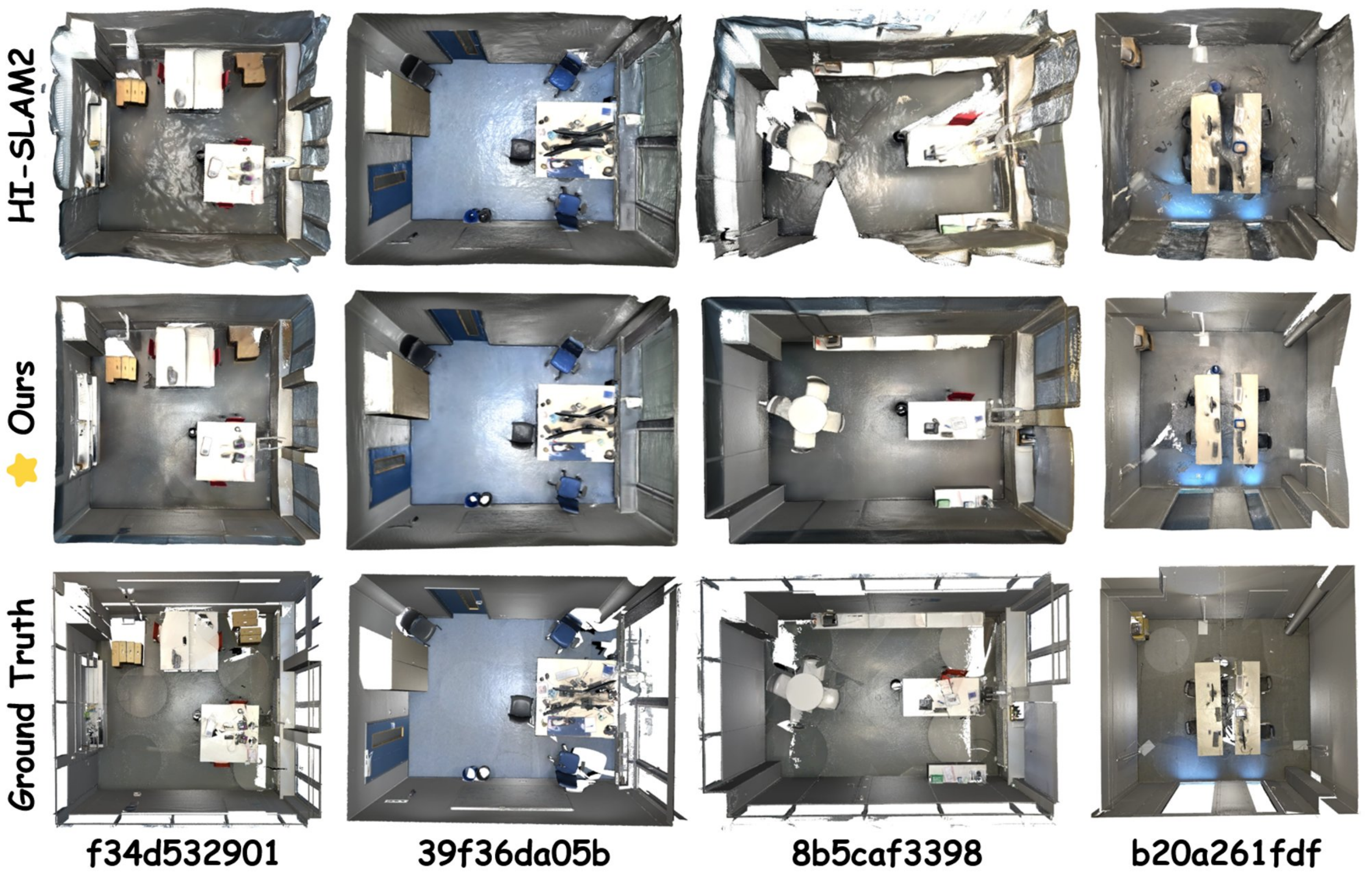}
\caption{Qualitative geometry reconstruction comparison between our method and representative baselines on the ScanNet++ dataset~\cite{Scannet++}. Our method achieves smoother surfaces.}
\label{fig:mesh_scannetpp}
\end{figure}

\noindent\textbf{Evaluation Metrics} We report the commonly used root mean square error (RMSE) of the absolute trajectory error (ATE) for camera tracking. Rendering quality is measured by PSNR, SSIM, and LPIPS on all input frames, with each method evaluated using its estimated poses. Following LoopSplat~\cite{loopsplat}, we report Depth L1 for 2D geometric quality and F1-score for 3D reconstruction. Since online SLAM systems must balance accuracy and efficiency, we also report FPS, GPU peak memory, and model size in the efficiency analysis. All experiments are run five times and the average values are reported. Failed runs are indicated where applicable. The best results are highlighted using: \firstbox{first}, \secondbox{second}, and \thirdbox{third}.

\noindent\textbf{Baselines} We compare DrawSplat against a wide range of recent SLAM systems, including Photo-SLAM~\cite{Photo-slam}, SplaTAM~\cite{Splatam}, MonoGS~\cite{MonoGS}, RTG-SLAM~\cite{Rtg-slam}, Splat-SLAM~\cite{splat-slam}, LoopSplat~\cite{loopsplat}, HI-SLAM2~\cite{HI-SLAM2}, MG-SLAM~\cite{MG-SLAM}, and SEGS-SLAM~\cite{Segs-slam}. For geometry reconstruction, we additionally include NeRF-based methods such as NICE-SLAM~\cite{Nice-slam}, Co-SLAM~\cite{Co-slam}, and ESLAM~\cite{Eslam}, since these methods remain strong references for geometric fidelity.

\begin{table}[t]
	\centering
	\caption{Quantitative comparison of appearance rendering performance on the Replica dataset~\cite{replica}.}
	\label{tab:rendering_replica}
	\renewcommand{\arraystretch}{1}
	\setlength{\tabcolsep}{1.5pt}
	\resizebox{0.498\textwidth}{!}{
		\begin{tabular}{llccccccccc}
			\toprule
			\textbf{Method} & \textbf{Metric} & Ro0 & Ro1 & Ro2 & Off0 & Off1 & Off2 & Off3 & Off4 & \textbf{Avg.} \\
			\midrule	
			\multirow{3}{*}{\makecell[l]{Spla\\TAM~\cite{Splatam}}} 
			& PSNR $\uparrow$ & 32.84 & 33.81 & 35.44 & 38.52 & 38.72 & 31.78 & 30.15 & 31.55 & 34.10 \\
			& SSIM $\uparrow$ & \third 0.968 & 0.965 & 0.970 & 0.981 & 0.975 & 0.955 & 0.943 & 0.941 & 0.962 \\
			& LPIPS $\downarrow$& 0.072 & 0.095 & 0.071 & 0.083 & 0.098 & 0.099 & 0.118 & 0.153 & 0.099 \\
			\midrule
			\multirow{3}{*}{\makecell[l]{Mono\\GS~\cite{MonoGS}} }
			& PSNR $\uparrow$ & 34.57 & 36.05 & 37.57 & 42.11 & 42.03 & 36.33 & 36.38 & 36.16 & 37.65 \\
			& SSIM $\uparrow$ & 0.961 & 0.952 & 0.973 & 0.980 & 0.976 & 0.972 & 0.969 & 0.965 & 0.969 \\
			& LPIPS $\downarrow$& 0.085 & 0.088 & 0.079 & 0.053 & 0.064 & 0.068 & 0.079 & 0.088 & 0.076 \\
			\midrule
			\multirow{3}{*}{\makecell[l]{RTG-\\SLAM~\cite{Rtg-slam}}} 
			& PSNR $\uparrow$ & 32.35 & 34.68 & 35.48 & 39.88 & 39.81 & 33.65 & 33.39 & 36.57 & 35.73 \\
			& SSIM $\uparrow$ & 0.960 & 0.977 & 0.980 & \second 0.989 & \second 0.989 & \second 0.980 & \second 0.981 & \second 0.984 & \second 0.980 \\
			& LPIPS $\downarrow$& 0.156 & 0.115 & 0.122 & 0.082 & 0.102 & 0.143 & 0.138 & 0.123 & 0.123 \\
			\midrule
			\multirow{3}{*}{\makecell[l]{Splat-\\SLAM~\cite{splat-slam}}} 
			& PSNR $\uparrow$ & 32.25 & 34.31 & 35.95 & 40.81 & 40.64 & 35.19 & 35.03 & 37.40 & 36.45 \\
			& SSIM $\uparrow$ & 0.910 & 0.930 & 0.950 & 0.980 & 0.970 & 0.960 & 0.950 & \third 0.980 & 0.954 \\
			& LPIPS $\downarrow$& 0.100 & 0.090 & 0.060 & 0.050 & \third 0.050 & 0.070 & 0.060 & \third 0.040 & 0.065 \\
			\midrule
			\multirow{3}{*}{\makecell[l]{Loop\\Splat~\cite{loopsplat}} }
			& PSNR $\uparrow$ & 33.08 & 35.12 & 36.09 & 40.87 & 40.18 & 34.71 & 35.51 & 37.08 & 36.58 \\
			& SSIM $\uparrow$ & \second 0.973 & \first 0.981 & \first 0.986 & \first 0.992 & \first 0.991 & \first 0.986 & \first 0.990 & \first 0.990 & \first 0.986 \\
			& LPIPS $\downarrow$& 0.118 & 0.118 & 0.112 & 0.088 & 0.130 & 0.142 & 0.095 & 0.101 & 0.113 \\
			\midrule
			\multirow{3}{*}{\makecell[l]{HI-\\SLAM2~\cite{HI-SLAM2}} }
			& PSNR $\uparrow$ & \third 35.06 & \third 36.45 & \third 38.24 & \second 42.41 & \second 43.54 & \third 36.73 & \third 36.83 & \second 38.45 & \third 38.46 \\
			& SSIM $\uparrow$ & 0.958 & 0.964 & 0.973 & 0.983 & 0.984 & 0.968 & 0.969 & 0.972 & 0.971 \\
			& LPIPS $\downarrow$& \third 0.042 & \third 0.047 & \third 0.036 & \third 0.021 & \first 0.021 & \third 0.042 & \third 0.037 & 0.043 & \third 0.036 \\
			\midrule
			\multirow{3}{*}{\makecell[l]{SEGS-\\SLAM~\cite{Segs-slam}} }
			& PSNR $\uparrow$ & \second 37.04 & \first 39.24 & \first 40.03 & \third 42.22 & \third 42.91 & \second 36.75 & \second 37.26 & \first 39.07 & \second 39.32 \\
			& SSIM $\uparrow$ & 0.966 & \third 0.978 & \third 0.980 & 0.981 & 0.979 & 0.969 & 0.967 & 0.975 & \third 0.974 \\
			& LPIPS $\downarrow$& \second 0.026 & \second 0.018 & \first 0.015 & \second 0.019 & \first 0.021 & \second 0.036 & \second 0.025 & \first 0.016 & \second 0.022 \\
			\midrule
			\multirow{3}{*}{\textbf{Ours}} 
			& PSNR $\uparrow$ & \first 37.43 & \second 39.12 & \second 38.35 & \first 42.54 & \first 43.84 & \first 37.89 & \first 37.92 & \third 38.35 & \first 39.43 \\
			& SSIM $\uparrow$ & \first 0.976 & \second 0.981 & \second 0.981 & \third 0.985 & \third 0.985 & \third 0.978 & \third 0.978 & 0.976 & \second 0.980 \\
			& LPIPS $\downarrow$& \first 0.018 & \first 0.016 & \second 0.019 & \first 0.015 & \second 0.025 & \first 0.023 & \first 0.019 & \second 0.027 & \first 0.020 \\
			
			\bottomrule
		\end{tabular}
	}
\end{table}

\begin{table}[t]
	\centering
	\caption{Quantitative comparison of appearance rendering performance on the ScanNet dataset~\cite{Scannet}.}
	\label{tab:rendering_scannet}
	\renewcommand{\arraystretch}{1}
	\setlength{\tabcolsep}{3.5pt}
	\resizebox{0.498\textwidth}{!}{
		\begin{tabular}{l l ccccccc}
			\toprule
			\textbf{Method} & \textbf{Metric}  & 0000 & 0059 & 0106 & 0169 & 0207 & \textbf{Avg.} \\
			\midrule
			
			\multirow{3}{*}{SplaTAM~\cite{Splatam}}
			& PSNR $\uparrow$& 17.95 & 20.11 & 19.85 & 20.09 & 17.98 & 19.20 \\
			& SSIM $\uparrow$& 0.611 & 0.796 & 0.739 & 0.795 & 0.710 & 0.730 \\
			& LPIPS $\downarrow$& 0.472 & 0.291 & 0.333 & 0.261 & 0.346 & 0.341 \\
			\midrule
			\multirow{3}{*}{MonoGS~\cite{MonoGS}}
			& PSNR $\uparrow$& 20.28 & 20.05 & 20.11 & 20.56 & 20.95 & 20.39 \\
			& SSIM $\uparrow$& 0.773 & 0.768 & 0.785 & 0.792 & 0.802 & 0.784 \\
			& LPIPS $\downarrow$& \third 0.385 & 0.295 & \third 0.273 & \third 0.251 & 0.293 & \third 0.299 \\
			\midrule
			\multirow{3}{*}{Photo-SLAM~\cite{Photo-slam}}
			& PSNR $\uparrow$& 20.16 & 20.25 & 19.31 & 20.34 & 20.32 & 20.08 \\
			& SSIM $\uparrow$& 0.763 & 0.801 & 0.758 & 0.779 & 0.768 & 0.774 \\
			& LPIPS $\downarrow$& 0.413 & \third 0.284 & 0.295 & 0.289 & 0.316 & 0.319 \\
			\midrule
			\multirow{3}{*}{RTG-SLAM~\cite{Rtg-slam}}
			& PSNR $\uparrow$& 18.75 & 16.02 & 15.21 & 18.05 & 18.88 & 17.38 \\
			& SSIM $\uparrow$& 0.758 & 0.686 & 0.733 & 0.777 & 0.779 & 0.747 \\
			& LPIPS $\downarrow$& 0.465 & 0.542 & 0.481 & 0.458 & 0.453 & 0.480 \\
			\midrule
			\multirow{3}{*}{MG-SLAM*~\cite{MG-SLAM}}
			& PSNR $\uparrow$& \second 25.69 & \second 23.62 & \third 22.95 & \third 22.86 & \third 23.42 & \third 23.71 \\
			& SSIM $\uparrow$& \third 0.846 & \third 0.838 & \third 0.829 & 0.836 & \third 0.840 & \third 0.838 \\
			& LPIPS $\downarrow$& \second 0.282 & \second 0.255 & \second 0.260 & \second 0.236 & \second 0.277 & \second 0.262 \\
			\midrule
			\multirow{3}{*}{LoopSplat~\cite{loopsplat}}
			& PSNR $\uparrow$& \third 25.17 & \third 23.25 & \second 23.39 & \second 24.43 & \second 26.42 & \second 24.53 \\
			& SSIM $\uparrow$& \second 0.850 & 0.835 & \second 0.844 & \third 0.855 & \second 0.859 & \second 0.849 \\
			& LPIPS $\downarrow$& 0.446 & 0.401 & 0.413 & 0.389 & 0.418 & 0.413 \\
			\midrule
			\multirow{3}{*}{SEGS-SLAM~\cite{Segs-slam}}
			& PSNR $\uparrow$& 20.21 & 21.29 & 16.25 & 21.60 & 21.46 & 20.16 \\
			& SSIM $\uparrow$& 0.809 & \first 0.859 & 0.704 & \second 0.866 & 0.827 & 0.813 \\
			& LPIPS $\downarrow$& 0.386 & 0.324 & 0.433 & 0.295 & \third 0.287 & 0.345 \\
			\midrule
			\multirow{3}{*}{\textbf{Ours}}
			& PSNR $\uparrow$& \first 27.05 & \first 25.15 & \first 26.39 & \first 28.24 & \first 28.51 & \first 27.07 \\
			& SSIM $\uparrow$& \first 0.869 & \second 0.852 & \first 0.897 & \first 0.895 & \first 0.865 & \first 0.876 \\
			& LPIPS $\downarrow$& \first 0.167 & \first 0.198 & \first 0.141 & \first 0.143 & \first 0.176 & \first 0.165 \\
			
			\bottomrule
		\end{tabular}
	}
\end{table}

\subsection{Camera Tracking Evaluation}
We first evaluate whether the proposed geometry-guided front-end improves camera pose estimation under low-texture, noisy, and large-scale indoor conditions.
Tables~\ref{tab:tracking_replica},~\ref{tab:tracking_scannet}, and~\ref{tab:tracking_apart} summarize the tracking results on Replica, ScanNet, and Replica Apartment. On Replica, the gap among strong methods is relatively small because the simulator provides ideal RGB-D observations with minimal sensor uncertainty. Nevertheless, DrawSplat remains competitive with the best-performing systems, indicating that the explicit geometry-guided tracking front-end does not sacrifice pose quality even in favorable conditions.

The advantage of the proposed method becomes more evident on ScanNet, where depth noise, motion blur, and real-world appearance variation make tracking more difficult. In this setting, DrawSplat consistently achieves better accuracy than frame-to-model Gaussian SLAM baselines. These results indicate that combining explicit geometric tracking, point-line joint constraints, and global refinement is more robust than relying primarily on photometric alignment.

We further evaluate robustness on Replica Apartment, which contains multi-room layouts, longer trajectories, and stronger drift accumulation. Methods such as SplaTAM and MonoGS tend to accumulate noticeable drift over long trajectories, while DrawSplat remains more stable. Even compared with point-line baselines such as MG-SLAM, our method achieves lower trajectory error, showing that the gain comes from the coupling among point correspondences, line geometry, gravity estimation, and global pose refinement rather than from line features alone.

\subsection{Appearance Rendering Evaluation}
We next evaluate whether the proposed representation and optimization strategy improve appearance rendering quality while preserving tracking robustness. Tables~\ref{tab:rendering_replica},~\ref{tab:rendering_scannet}, and~\ref{tab:rendering_apart} report rendering performance on Replica, ScanNet, and Replica Apartment. On Replica, DrawSplat achieves consistently strong rendering quality while maintaining superior geometry quality. In particular, the proposed method performs favorably on PSNR and LPIPS, indicating that the structure-aware implicit Gaussian representation improves both pixel-level fidelity and perceptual consistency.

The gain is more pronounced on ScanNet, where sensor noise and uncertain depth often cause structural degradation in Gaussian SLAM maps. Geometry-guided tracking stabilizes Gaussian anchors, while structural regularization suppresses local shape distortions that would otherwise appear as rendering artifacts. The qualitative comparison results are shown in Fig.~\ref{fig:rendering}. On Replica Apartment, DrawSplat maintains high rendering quality despite large scenes and long trajectories, suggesting that the system preserves both trajectory consistency and scene coherence over long-range reconstruction.

Overall, the rendering results indicate that geometry-aware optimization improves appearance quality without degrading reconstruction consistency.

\subsection{Geometry Reconstruction Evaluation}
We evaluate geometric reconstruction quality in Table~\ref{tab:geometry_replica} and Table~\ref{tab:geometry_scannet}. On Replica, DrawSplat achieves the best or near-best Depth L1 results across most sequences, suggesting that the proposed representation uses depth supervision more effectively than conventional explicit Gaussian formulations. Our results on the real-world ScanNet dataset further demonstrate robustness to noisy depth inputs.

For 3D geometry reconstruction, we compare against both Gaussian-based and NeRF-based methods using the mesh-level evaluation protocol on Replica. The proposed method achieves the best reconstruction quality among Gaussian-based SLAM approaches and remains competitive with the strongest NeRF-based systems. In particular, DrawSplat attains the highest F1-score among Gaussian SLAM methods, reflecting the benefit of combining continuous representation with structural regularization.

As shown in Fig.~\ref{fig:mesh_scannet} and~\ref{fig:mesh_scannetpp}, the qualitative comparisons on ScanNet and ScanNet++ further support this conclusion. Relative to LoopSplat~\cite{loopsplat} and HI-SLAM2~\cite{HI-SLAM2}, DrawSplat reconstructs smoother low-texture surfaces, cleaner wall-floor intersections, and challenging structures such as transparent windows and large planar surfaces. These observations show that the proposed representation supports accurate geometry reconstruction rather than only rendering-oriented fitting.

\begin{table}[!t]
	\centering
	\caption{Quantitative comparison of appearance rendering performance on the Replica Apartment dataset~\cite{replica,Tandem}. * represents the baseline results of~\cite{Splatam,Photo-slam,MonoGS,Rtg-slam,MG-SLAM} come from MG-SLAM~\cite{MG-SLAM}. The dash indicates system failures.}
	\label{tab:rendering_apart}
	\setlength{\tabcolsep}{3pt}
	\resizebox{0.498\textwidth}{!}{
		\begin{tabular}{llcccccc}
			\toprule
			\textbf{Method} & \textbf{Metric} & apart0 & apart1 & apart2 & frl0 & frl4 & \textbf{Avg.} \\
			\midrule
			
			\multirow{3}{*}{SplaTAM*~\cite{Splatam}}
			& PSNR $\uparrow$  & 13.12 & 24.57 & 28.52 & 31.82 & 32.71 & 26.15 \\
			& SSIM $\uparrow$  & 0.415 & 0.821 & 0.883 & 0.930 & 0.945 & 0.799 \\
			& LPIPS $\downarrow$ & 0.656 & 0.302 & 0.241 & 0.164 & 0.184 & 0.309 \\
			\midrule
			\multirow{3}{*}{Photo-SLAM*~\cite{Photo-slam}}
			& PSNR $\uparrow$  & 29.07 & 22.73 & 24.59 & 34.16 & 33.36 & 28.78 \\
			& SSIM $\uparrow$  & 0.922 & 0.796 & 0.848 & 0.940 & 0.932 & 0.888 \\
			& LPIPS $\downarrow$ & 0.227 & 0.293 & 0.354 & 0.115 & 0.129 & 0.224 \\
			\midrule
			\multirow{3}{*}{MonoGS*~\cite{MonoGS}}
			& PSNR $\uparrow$  & 22.91 & 26.88 & 27.93 & 31.72 & 27.98 & 27.48 \\
			& SSIM $\uparrow$  & 0.864 & 0.835 & 0.836 & 0.886 & 0.873 & 0.859 \\
			& LPIPS $\downarrow$ & 0.385 & 0.284 & 0.272 & 0.225 & 0.245 & 0.282 \\
			\midrule
			\multirow{3}{*}{RTG-SLAM*~\cite{Rtg-slam}}
			& PSNR $\uparrow$  & - & \third{29.08} & 29.14 & 33.88 & - & - \\
			& SSIM $\uparrow$  & - & \third{0.900} & 0.909 & 0.933 & - & - \\
			& LPIPS $\downarrow$ & - & \second{0.232} & \third{0.232} & 0.181 & - & - \\
			\midrule
			\multirow{3}{*}{MG-SLAM*~\cite{MG-SLAM}}
			& PSNR $\uparrow$  & \third{34.54} & \second{32.37} & \third{32.92} & \third{36.24} & \third{35.66} & \third{34.35} \\
			& SSIM $\uparrow$  & \third{0.970} & \second{0.925} & \third{0.946} & \third{0.966} & \third{0.972} & \third{0.956} \\
			& LPIPS $\downarrow$ & \third{0.269} & \third{0.255} & 0.244 & \third{0.168} & \third{0.180} & \third{0.223} \\
			\midrule
			\multirow{3}{*}{LoopSplat~\cite{loopsplat}}
			& PSNR $\uparrow$  & \second{37.53} & 27.97 & \second{33.71} & \second{36.66} & \second{37.31} & \second{34.64} \\
			& SSIM $\uparrow$  & \first{0.982} & 0.899 & \second{0.975} & \first{0.986} & \first{0.984} & \second{0.965} \\
			& LPIPS $\downarrow$ & \second{0.131} & 0.296 & \second{0.175} & \second{0.108} & \second{0.133} & \second{0.169} \\
			\midrule
			\multirow{3}{*}{\textbf{Ours}}
			& PSNR $\uparrow$  & \first{38.49} & \first{37.35} & \first{38.22} & \first{40.30} & \first{39.60} & \first{38.79} \\
			& SSIM $\uparrow$  & \second{0.973} & \first{0.972} & \first{0.976} & \second{0.980} & \second{0.974} & \first{0.975} \\
			& LPIPS $\downarrow$ & \first{0.043} & \first{0.030} & \first{0.028} & \first{0.018} & \first{0.028} & \first{0.029} \\
			
			\bottomrule
		\end{tabular}
	}
\end{table}

\subsection{Performance Evaluation}
Since visual SLAM systems must balance accuracy and online efficiency, we additionally analyze computational cost, rendering speed, and memory footprint. Table~\ref{tab:performance} compares DrawSplat with representative baselines in terms of tracking FPS, rendering FPS, SLAM FPS, GPU peak memory, and model size.  SLAM FPS is computed as the total number of input frames divided by the wall-clock time required to complete SLAM processing for the sequence, including tracking, mapping, plane estimation, loop closure, and bundle adjustment. Semantic preprocessing is timed separately and reported in Table~\ref{tab:ablation_semantic_source}. DrawSplat achieves real-time tracking and fast rendering while providing better reconstruction quality than lightweight Gaussian baselines.

For mapping, instead of using conventional explicit Gaussian primitives as in Photo-SLAM~\cite{Photo-slam} and MonoGS~\cite{MonoGS}, DrawSplat adopts a voxel-organized implicit Gaussian representation to improve reconstruction quality. This representation introduces additional decoding and mapping costs, leading to a lower rendering speed than highly optimized explicit Gaussian systems. Nevertheless, DrawSplat achieves the best rendering and geometric quality with the lowest GPU memory and the smallest model size. Compared with SEGS-SLAM~\cite{Segs-slam}, which also uses an implicit Gaussian representation, DrawSplat maintains high-fidelity rendering while exploiting the Atlanta World assumption and adaptive structural sampling to reduce the model size to less than 20 MB.

\section{Ablation Study}

\subsection{Ablation on the Tracking Front-end}
We first study which components of the explicit geometry-guided tracking front-end are responsible for the observed gains in pose accuracy and downstream reconstruction quality. Table~\ref{tab:ablation_tracking_frontend} presents the performance of different variants. The tracking ablation shows that point-only tracking is prone to drift in weak-texture and repetitive structural regions. Adding line features reduces ATE, confirming that line segments provide more reliable constraints in wall corners, door frames, corridors, and staircases. Adding global BA on top of point-only tracking improves long-range consistency, but still does not match the accuracy of point-line joint tracking, indicating that stronger local geometric constraints cannot be fully replaced by back-end correction alone.

\begin{table}[t]
	\centering
\caption{Quantitative comparison of geometry reconstruction accuracy on the Replica dataset~\cite{replica}.}
	\label{tab:geometry_replica}
	\renewcommand{\arraystretch}{1}
	\setlength{\tabcolsep}{1.5pt}
	\resizebox{0.498\textwidth}{!}{
		\begin{tabular}{llccccccccc}
			\toprule
			\textbf{Method} & \textbf{Metric} & Ro0 & Ro1 & Ro2 & Off0 & Off1 & Off2 & Off3 & Off4 & \textbf{Avg.} \\
			\midrule
			
			\rowcolor{gray!10}
			\multicolumn{11}{l}{\textbf{Neural Implicit Fields}} \\
			
			\multirow{2}{*}{NICE-SLAM~\cite{Nice-slam}} 
			& Depth L1 $\downarrow$& 1.75 & 1.34 & 1.99 & 1.45 & 1.65 & 6.95 & 4.82 & 2.31 & 2.78 \\
			& F1-score $\uparrow$& 90.2 & 91.9 & 92.2 & 93.1 & 82.2 & 88.0 & 87.1 & 87.7 & 89.1 \\
			\midrule
			Co-SLAM~\cite{Co-slam} 
			& Depth L1 $\downarrow$& 1.05 & 0.85 & 2.37 & 1.24 & 1.48 & 1.86 & 1.66 & 1.54 & 1.51 \\
			\midrule
			\multirow{2}{*}{ESLAM~\cite{Eslam}} 
			& Depth L1 $\downarrow$& 0.99 & 1.21 & 1.17 & 0.97 & 1.19 & 1.62 & \third 1.55 & 1.01 & 1.21 \\
			& F1-score $\uparrow$& \first 94.8 & \first 95.0 & \first 95.7 & \first 97.5 & \first 95.8 & \first 93.1 & \first 93.9 & \first 92.2 & \first 94.7 \\
			
			\midrule
			
			\rowcolor{gray!10}
			\multicolumn{11}{l}{\textbf{3D Gaussian Splatting}} \\
			
			\multirow{2}{*}{SplaTAM~\cite{Splatam}} 
			& Depth L1 $\downarrow$& \second 0.48 & \third 0.41 & \third 0.56 & \second 0.45 & \third 0.70 & \third 1.05 & 1.57 & \third 0.69 & \third 0.74 \\
			& F1-score $\uparrow$& 89.1 & 89.8 & 87.7 & 92.3 & 90.2 & 86.9 & 79.9 & 81.3 & 87.1 \\
			\midrule
			\multirow{2}{*}{MonoGS~\cite{MonoGS}} 
			& Depth L1 $\downarrow$& 0.79 & 0.56 & 0.91 & \third 0.70 & 1.01 & \second 0.95 & \second 0.97 & 1.66 & 0.94 \\
			& F1-score $\uparrow$& 80.2 & 83.5 & 81.0 & 85.3 & 84.2 & 79.1 & 79.9 & 59.5 & 79.1 \\
			\midrule
			\multirow{2}{*}{RTG-SLAM~\cite{Rtg-slam}} 
			& Depth L1 $\downarrow$& 2.41 & 1.13 & 2.05 & 0.87 & 0.76 & 1.70 & 2.66 & 2.05 & 1.70 \\
			& F1-score $\uparrow$& 91.5 & 92.4 & 90.6 & 87.1 & 85.9 & 87.9 & 88.4 & 89.5 & 89.2 \\
			\midrule
			\multirow{2}{*}{LoopSplat~\cite{loopsplat}} 
			& Depth L1 $\downarrow$& \first 0.44 & \first 0.28 & \first 0.40 & \second 0.45 & \first 0.21 & 1.32 & 2.39 & \first 0.35 & \second 0.73 \\
			& F1-score $\uparrow$& \second 92.2 & \second 93.2 & \second 93.0 & \second 94.9 & \second 92.5 & \second 90.8 & \second 90.8 & \second 90.4 & \second 92.2 \\
			\midrule
			HI-SLAM2~\cite{HI-SLAM2} 
			& F1-score $\uparrow$& 90.9 & 91.2 & 88.9 & 91.3 & 90.5 & 86.0 & 87.1 & 87.2 & 89.1 \\
			\midrule
			\multirow{2}{*}{\textbf{Ours}} 
			& Depth L1 $\downarrow$& \third 0.64 & \second 0.36 & \second 0.47 & \first 0.39 & \second 0.36 & \first 0.66 & \first 0.77 & \second 0.53 & \first 0.52 \\
			& F1-score $\uparrow$& \second 94.0 & \second 94.2 & \second 94.6 & \second 96.9 & \second 94.4 & \second 91.8 & \second 92.8 & \second 91.0 & \second 93.7 \\
			
			\bottomrule
		\end{tabular}
	}
\end{table}

\begin{table}[htbp]
	\centering
	\caption{Quantitative comparison of 2D geometry reconstruction accuracy (Depth L1 $\downarrow$) on the ScanNet dataset~\cite{Scannet}.}
	\label{tab:geometry_scannet}
	\resizebox{0.498\textwidth}{!}{
	\begin{tabular}{lcccccc}
		\toprule
		\textbf{Method} & 0000 & 0059 & 0106 & 0169 & 0207 & \textbf{Avg.} \\
		\midrule
		SplaTAM~\cite{Splatam}    & 7.87  & 5.63  & 7.13  & 5.28  & 6.40  & 6.46  \\
		MonoGS~\cite{MonoGS}     & 56.67 & 50.95 & 47.47 & 28.10 & 54.41 & 47.52 \\
		Splat-SLAM~\cite{splat-slam} & 8.21  & 15.54 & 18.43 & 11.31 & 10.54 & 12.81 \\
		LoopSplat\cite{loopsplat}  & 10.93 & 25.16 & 45.42 & 45.72 & 17.18 & 28.88 \\
		\textbf{Ours} & \first{3.05} & \first{5.06} & \first{4.76} & \first{5.26} & \first{3.91} & \first{4.41} \\
		\bottomrule
	\end{tabular}
	}
\end{table}

\begin{table}[!tbp]
	\centering
	\renewcommand{\arraystretch}{1}
	\setlength{\tabcolsep}{3.5pt}
\caption{System performance comparison. The reported values are averaged over all sequences in the Replica dataset~\cite{replica}. Experiments are conducted on an NVIDIA A100 GPU and an AMD EPYC 7542 CPU. All images are processed at their original resolution. DrawSplat achieves strong reconstruction quality with a compact model, real-time tracking, and fast rendering.}
	\label{tab:performance}
	\resizebox{0.498\textwidth}{!}{
		\begin{tabular}{l| c c c c c c}
			\toprule
			\textbf{Method} & 
			\begin{tabular}[c]{@{}c@{}}{PSNR}\\ {[dB]} $\uparrow$\end{tabular} &
			\begin{tabular}[c]{@{}c@{}}{Track.}\\ {FPS} $\uparrow$\end{tabular} & 
			\begin{tabular}[c]{@{}c@{}}{Render.}\\ {FPS} $\uparrow$\end{tabular} & 
			\begin{tabular}[c]{@{}c@{}}{SLAM}\\ {FPS} $\uparrow$\end{tabular} & 
			\begin{tabular}[c]{@{}c@{}}{Memory}\\ {[GB]} $\downarrow$\end{tabular} & 
			\begin{tabular}[c]{@{}c@{}}{Size}\\ {[MB]} $\downarrow$\end{tabular} \\
			\midrule
			SplaTAM \cite{Splatam}        & 34.10        & 0.26        & 121          & 0.13           & 18.8          & 295.5      \\
			Photo-SLAM \cite{Photo-slam}    & 34.86     & \first{28.87}      & \first{494}    & \first{13.02}   & \third{7.3}     & \third{61.75}     \\
			MonoGS  \cite{MonoGS}     & \third{37.65}     & 2.11     & \second{465}    & 1.28     & 16.5      & \second{44.23}             \\
			LoopSplat \cite{loopsplat}      & 36.58      & 0.68       & 138       & 0.35     & 9.8                     & 73.28                   \\
			SEGS-SLAM  \cite{Segs-slam}       & \second{39.32}   & \third{15.29}    & 143     & \third{1.73}    & \second{4.7}     & 78.98      \\
			\textbf{Ours}     & \first{39.43}     & \second{24.55}      & \third{209}      & \second{2.02}       & \first{4.5}   & \first{18.29} \\
			\bottomrule         
		\end{tabular}
	}
\end{table}

Rows (e) and (f) both use point-induced planes, with point-based and line-based gravity, respectively; row (g) uses line-induced planes and line-based gravity. Comparing (e) with (f) isolates the gravity source: line-based gravity slightly improves PSNR on both sequences without changing ATE. Comparing (f) with (g) isolates the plane source: line-induced planes further improve PSNR and tracking FPS on both sequences, while ATE remains unchanged.

\subsection{Ablation on the Representation Design}

\begin{table}[!t]
	\centering
	\caption{Ablation study of the tracking front-end on Replica Apartment~\cite{replica,Tandem} apart0 and ScanNet~\cite{Scannet} scene0000. ATE RMSE $\downarrow$ is reported in [cm].}
	\label{tab:ablation_tracking_frontend}
	\renewcommand{\arraystretch}{1}
	\setlength{\tabcolsep}{1pt}
	\resizebox{0.498\textwidth}{!}{
		\begin{tabular}{l|l|ccc|ccc}
			\toprule
			& \multirow{2}{*}{\textbf{Method}} 
			& \multicolumn{3}{|c|}{{apart0}} 
			& \multicolumn{3}{c}{{scene0000}} \\
			\cmidrule(lr){3-5}\cmidrule(lr){6-8}
			& 
			& ATE $\downarrow$ 
			& PSNR $\uparrow$ 
			& \begin{tabular}[c]{@{}c@{}}Track.\\ FPS $\uparrow$\end{tabular}
			& ATE $\downarrow$ 
			& PSNR $\uparrow$ 
			& \begin{tabular}[c]{@{}c@{}}Track.\\ FPS $\uparrow$\end{tabular} \\
			\midrule
			(a) & Only point                    
			& 8.12          & 34.86          & \textbf{34.72} 
			& 6.25          & 24.18          & \textbf{41.36} \\
			(b) & (a) + Line               
			& 3.81  & 37.42          & 25.64 
			& 5.48  & 26.31          & 31.75 \\
			(c) & (a) + Global BA          
			& 5.35   & 36.05          & 30.18 
			& 5.72   & 25.09          & 36.42 \\
			(d) & (b) + Global BA  
			& \textbf{3.21}   & 38.21  & 23.16 
			& \textbf{5.03}   & 26.86  & 29.37 \\
			\midrule 
			(e) & (d) + Point planes + point gravity
			& \textbf{3.21}   & 38.03   & 21.74 
			& \textbf{5.03}   & 26.54   & 26.83 \\
			(f) & (d) + Point planes + line gravity
			& \textbf{3.21}   & 38.11   & 21.59 
			& \textbf{5.03}   & 26.75   & 26.76 \\
			(g) & \textbf{Ours}                     
			& \textbf{3.21}   & \textbf{38.49}   & 22.58 
			& \textbf{5.03}   & \textbf{27.05}   & 28.64 \\
			\bottomrule
		\end{tabular}
	}
\end{table}

\begin{table}[!t]
	\centering
	\renewcommand{\arraystretch}{1}
	\setlength{\tabcolsep}{1.2pt}
	\caption{Ablation study of the representation design. The reported values are averaged over all sequences in the Replica dataset~\cite{replica}. * represents that we further add depth optimization to GOF~\cite{gof}.}
	\label{tab:ablation_representation}
	\resizebox{0.498\textwidth}{!}{
		\begin{tabular}{l|ccccc}
			\toprule
			Method   & PSNR $\uparrow$      & F1-score $\uparrow$            & Render. FPS $\uparrow$ & Gaussians $\downarrow$ & Size [MB] $\downarrow$\\
			\midrule
			Our Track. + 3DGS~\cite{3dgs} & \textbf{40.57} & - & \textbf{346} & 437231 & 108.43\\
			Our Track. + GOF*~\cite{gof} & {39.05} & 89.57 & {279} & 1036097 & 261.10\\
			\midrule
			voxel\_size = 0.001       & 40.48            & 92.66          & 152 & 127712 & 54.15           \\
			voxel\_size = 0.01 (\textbf{ours}) & 39.43   & \textbf{93.71} & 209 & 53380 & 18.29           \\
			voxel\_size = 0.1         & 35.92                    & 92.62          & 238 & \textbf{6529} & \textbf{2.77}  \\
			\bottomrule 
		\end{tabular}
	}
\end{table}

\begin{figure}[!tbp]
	\centering
	\includegraphics[width=0.98\linewidth]{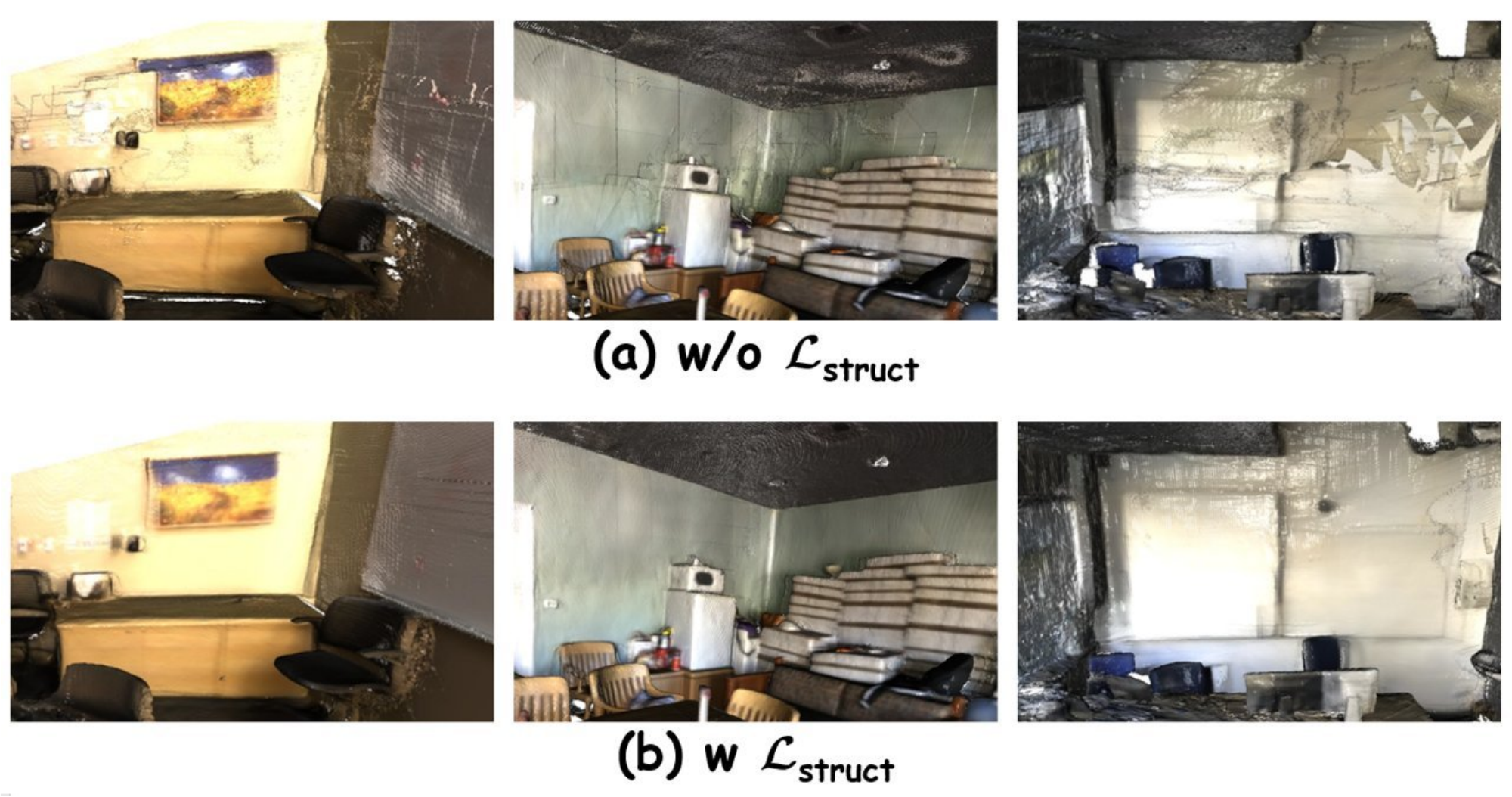}
	\caption{Ablation study of  Atlanta World-guided structural regularization on ScanNet dataset~\cite{Scannet}. The proposed regularization reduces sensor noise and produces smoother geometric structures.}
	\label{fig:ablation_struct}
\end{figure}

As shown in Table~\ref{tab:ablation_representation}, 
We first compare the proposed implicit Gaussian representation with conventional explicit 3DGS~\cite{3dgs} and the GOF variant~\cite{gof} while keeping the tracking front-end unchanged. The explicit 3DGS baseline achieves higher PSNR and rendering speed, but produces a substantially larger Gaussian point cloud. GOF provides geometric reconstruction, but also requires a much larger model and yields a lower F1-score than DrawSplat.

The remaining rows evaluate the voxel-size sensitivity of the proposed implicit representation. A small voxel size increases representation freedom but introduces redundant Gaussians, increasing the model size from 18.29 MB to 54.15 MB and reducing rendering speed. A large voxel size reduces the number of Gaussians and model size but degrades local detail and surface precision. The default voxel size provides a favorable balance among reconstruction quality, efficiency, and compactness.

\subsection{Ablation on Adaptive Structural Sampling}

We then evaluate the contribution and characterize the quality--efficiency trade-off of adaptive structural sampling. As shown in Table~\ref{tab:ablation_sampling}, removing structure-aware pruning preserves more low-opacity or low-importance anchors and can slightly improve some metrics, but at a clear computational cost. In contrast, the full strategy removes only sufficiently observed anchors with both low opacity and low structural importance, improving efficiency while preserving reconstruction quality. 

Removing the line response, depth-gradient response, or photometric residual reduces both PSNR and F1-score. These variants have model sizes of 17.60--18.11 MB, compared with 18.29 MB for the full method. The results support the contribution of each response within the complete sampling strategy and characterize the accompanying changes in model size and rendering speed.

When the unified structural weight $w(\mathbf{x})$ is removed, densification degenerates toward a conventional gradient-driven policy. In this case, the model can no longer explicitly focus on high-information regions such as wall corners, door frames, or depth discontinuities, and both rendering and reconstruction quality degrade. If the method is further simplified to a purely gradient-based densification strategy, the reconstruction quality drops even more, despite the higher rendering speed.

Overall, adaptive structural sampling reallocates capacity toward geometrically difficult regions and improves the quality-efficiency trade-off.

\subsection{Ablation on Semantic Source}

We further evaluate robustness to different semantic sources. Since semantic cues identify structural regions and modulate plane-to-Gaussian associations, directly removing semantics would change the intended weighting mechanism. We therefore compare ground-truth semantic labels with 2D predicted semantic maps. As shown in Table~\ref{tab:ablation_semantic_source}, predicted semantics achieves nearly identical PSNR and F1-score to ground-truth semantics~\cite{SemanticNeRF}, with high IoU scores for the major structural classes. The results indicate that DrawSplat does not rely on ground-truth semantic labels and remains compatible with practical 2D semantic predictions. We also report the preprocessing speed for the predicted semantic maps.

\subsection{Ablation on the Loss Design}

\begin{table}[!t]
	\centering
	\renewcommand{\arraystretch}{1}
	\setlength{\tabcolsep}{1.2pt}
	\caption{Ablation study of the proposed adaptive structural sampling strategy on the Replica dataset~\cite{replica}.}
	\label{tab:ablation_sampling}
	\resizebox{0.498\textwidth}{!}{
		\begin{tabular}{l|cccc}
			\toprule
			\textbf{Method}
			& PSNR $\uparrow$
			& F1-score $\uparrow$
			& Size [MB] $\downarrow$
			& Render. FPS $\uparrow$ \\
			\midrule
			Gradient-based densification
			& 38.61 & 91.92 & 16.95 & 217 \\
			
			w/o structural weight
			& 39.02 & 92.76 & 16.82 & 218 \\
			
			w/o line response
			& 39.12	& 92.87 & 17.60 & 212 \\
			
			w/o depth-gradient response	& 39.21 & 93.10 & 17.74 & 211 \\
			
			w/o photometric residual & 39.13 & 93.21 & 18.11 & 213 \\
			
			w/o pruning
			& {39.55} & {93.84} & 23.17 & 166 \\
			
			\textbf{Ours}
			& 39.43 & 93.71 & 18.29 & 209 \\
			\bottomrule
		\end{tabular}
	}
\end{table}

\begin{table}[!t]
	\centering
	\caption{Ablation study of semantic source robustness on the Replica dataset~\cite{replica}.}
	\label{tab:ablation_semantic_source}
	\renewcommand{\arraystretch}{1}
	\setlength{\tabcolsep}{1.5pt}
	\resizebox{0.498\textwidth}{!}{
		\begin{tabular}{l|cccccc}
			\toprule
			{Source} & PSNR $\uparrow$  & F1-score $\uparrow$ & mIoU$_w$ $\uparrow$ & IoU$_f$ $\uparrow$ & IoU$_c$ $\uparrow$  & FPS $\uparrow$ \\
			\midrule
			GT sem. & {39.48} & {93.81} & {0.989} & 0.954 & {0.836} & - \\
			Predicted sem. & 39.43 & 93.71 & 0.986 & {0.964} & 0.809 & 9.2\\
			\bottomrule
		\end{tabular}
	}
\end{table}


\begin{table}[!t]
	\centering
	\renewcommand{\arraystretch}{1}
	\setlength{\tabcolsep}{3.5pt}
	\caption{Ablation study of the map loss and semantic weighting design on the Replica dataset~\cite{replica}.}
	\label{tab:ablation_loss}
	\resizebox{0.498\textwidth}{!}{
		\begin{tabular}{l|ccc}
			\toprule
			\textbf{Method}
			& PSNR $\uparrow$
			& Depth L1 $\downarrow$
			& F1-score $\uparrow$ \\
			\midrule
			Basic
			& \textbf{39.85}
			& 3.86
			& 73.39 \\
			\midrule
			w/o $\mathcal{L}_{d}$
			& 38.02
			& 2.97
			& 77.22 \\
			
			w/o $\mathcal{L}_{n}$
			& 39.23
			& 0.68
			& 93.44 \\
			\midrule
			w/o $\mathcal{L}_{\mathrm{gp}}$
			& 39.08
			& 0.64
			& 93.31 \\
			
			w/o $\mathcal{L}_{\mathrm{gn}}$
			& 39.25
			& 0.58
			& 93.50 \\
			
			w/o $\mathcal{L}_{\mathrm{aw}}$
			& 39.33
			& 0.54
			& 93.55 \\
			
			w/o $\mathcal{L}_{\mathrm{struct}}$
			& 37.55
			& 0.55
			& 93.23 \\
			\midrule
			w/o semantic weighting
			& 39.32
			& 0.57
			& 93.45 \\
			\midrule
			\textbf{Ours}
			& {39.43}
			& \textbf{0.52}
			& \textbf{93.71} \\
			\bottomrule
		\end{tabular}
	}
\end{table}

Finally, we analyze the contribution of the major losses and semantic weighting design in the map optimization objective and report the performance of different variants in Table~\ref{tab:ablation_loss}. Using only the appearance loss yields high PSNR but poor geometry, showing that image fitting alone is insufficient for high-quality scene reconstruction. Removing depth supervision degrades geometric performance, confirming that depth is essential for stabilizing surface positions and suppressing global drift. Removing the normal loss leads to weaker local surface smoothness, indicating that normal consistency is important for improving planar regularity.

Removing the structural loss reduces both geometry quality and reconstruction stability. On Replica, it weakens planar consistency and boundary sharpness; on ScanNet, Fig.~\ref{fig:ablation_struct} shows that the structural loss is important for preserving global coherence and mesh usability under severe sensor noise. The full objective therefore combines complementary losses, with the structural term helping align the Gaussian map with scene-level geometry. Specifically, removing $\mathcal{L}_{\mathrm{gp}}$ causes the largest degradation among the individual structural terms. This result indicates that explicitly constraining Gaussian centers toward structural planes is important for geometric reconstruction. Removing $\mathcal{L}_{\mathrm{gn}}$ or $\mathcal{L}_{\mathrm{aw}}$ also degrades the reconstruction metrics, confirming their complementary roles in normal alignment and Atlanta World orientation consistency.

The ``w/o semantic weighting'' variant sets $s_{kj}=1$ while retaining the structural planes and geometric distance terms. Its lower PSNR and F1-score and higher Depth L1 show that semantic compatibility improves plane-to-Gaussian association. Removing the full structural loss causes a larger drop in PSNR and F1-score, while the Basic variant confirms that appearance supervision alone is insufficient for geometric reconstruction.

\section{Conclusion}

We presented DrawSplat, an Atlanta World-guided implicit Gaussian SLAM framework for RGB-D indoor reconstruction. DrawSplat uses explicit point-line geometry to estimate stable poses, gravity cues, and structural planes, and converts these cues into differentiable regularization terms for an anchor-based implicit Gaussian map. Together with adaptive structural sampling, this design improves the balance among tracking accuracy, rendering quality, geometric reconstruction, and model compactness. Experiments on synthetic and real-world indoor datasets validate the complementary roles of explicit geometric tracking, structural regularization, implicit Gaussian representation, and adaptive sampling.

\noindent\textbf{Limitation and Future Work} DrawSplat is mainly designed for RGB-D indoor scenes with dominant architectural structures. Its performance may be less stable in highly dynamic environments, outdoor scenes, or spaces that weakly satisfy the Atlanta World assumption. Severe depth errors or unreliable lines may impair pose, gravity, and plane estimation. DrawSplat also cannot predict all the unobserved holes in the same way as the generation method~\cite{Gen3r} does.  Future work will explore more general layouts and monocular SLAM using metric depth prediction.



\bibliographystyle{IEEEtran}

\bibliography{reference}


 







\vfill

\end{document}